%% file: main.tex
\documentclass[letterpaper]{article} 
\usepackage{aaai2026}  
\usepackage{times}  
\usepackage{helvet}  
\usepackage{courier}  
\usepackage[hyphens]{url}  
\usepackage{graphicx} 
\usepackage{natbib}  
\usepackage{caption} 
\usepackage{algorithm}
\usepackage{algorithmic}

\usepackage{newfloat}
\usepackage{listings}

\usepackage{algorithm}
\usepackage{algorithmic}
\usepackage{subfigure}
\usepackage{booktabs} 
\usepackage{tcolorbox}
\usepackage{amsmath}
\usepackage{mathtools}
\usepackage{amsthm}
\usepackage{cancel}
\usepackage{pifont}
\usepackage{utfsym}
\input{math_commands}

\input{macros}

\DeclareCaptionStyle{ruled}{labelfont=normalfont,labelsep=colon,strut=off} 
\floatstyle{ruled}
\newfloat{listing}{tb}{lst}{}
\floatname{listing}{Listing}
\title{Variation-Bounded Loss for Noise-Tolerant Learning}
\author{
    Jialiang Wang\textsuperscript{\rm 1,2}\equalcontrib,
    Xiong Zhou\textsuperscript{\rm 1}\equalcontrib,
    Xianming Liu\textsuperscript{\rm 1},
    Gangfeng Hu\textsuperscript{\rm 1} \\
    Deming Zhai\textsuperscript{\rm 1}\thanks{Corresponding author.},
    Junjun Jiang\textsuperscript{\rm 1},
    Haoliang Li\textsuperscript{\rm 2}
}
\affiliations{
    \textsuperscript{\rm 1}Harbin Institute of Technology \quad \textsuperscript{\rm 2}City University of Hong Kong

}

\usepackage{bibentry}

\begin{document}

\maketitle

\newtheorem{Theorem}{Theorem} 
\newtheorem{Definition}{Definition}
\newtheorem{Proposition}{Proposition}
\newtheorem{Lemma}{Lemma}
\newtheorem{Corollary}{Corollary}
\renewcommand{\theequation}{\arabic{equation}}

\input{0_abstract}

\input{1_introduction}
\input{2_preliminary}
\input{3_methodology}

\input{4_experiments}
\input{5_conclusion}

\section{Acknowledgements}
This work was supported in part by National Key Research and Development Program of China under Grant 2023YFC2509100, in part by National Natural Science Foundation of China under Grants 62525107 and 632B2031,  and in part by Fundamental Research Funds for the Central Universities (HIT.DZJJ.2025055).

\bibliography{aaai2026}
\input{6_appendix}
\end{document}

%% file: math_commands.tex
\usepackage{amsmath,amsfonts,bm}

\def\eqref#1{equation~\ref{#1}}

\def\1{\bm{1}}

\def\rve{{\mathbf{e}}}

\def\rvu{{\mathbf{i}}}

\def\rvu{{\mathbf{u}}}
\def\rvv{{\mathbf{v}}}

\def\rvx{{\mathbf{x}}}

\DeclareMathAlphabet{\mathsfit}{\encodingdefault}{\sfdefault}{m}{sl}
\SetMathAlphabet{\mathsfit}{bold}{\encodingdefault}{\sfdefault}{bx}{n}

\def\gD{{\mathcal{D}}}

\def\gF{{\mathcal{F}}}

\def\gL{{\mathcal{L}}}

\def\gR{{\mathcal{R}}}
\def\gS{{\mathcal{S}}}

\def\gU{{\mathcal{U}}}

\def\gX{{\mathcal{X}}}
\def\gY{{\mathcal{Y}}}

\def\sR{{\mathbb{R}}}

\newcommand{\E}{\mathbb{E}}

\newcommand{\R}{\mathbb{R}}

\DeclareMathOperator*{\argmax}{arg\,max}
\DeclareMathOperator*{\argmin}{arg\,min}

%% file: macros.tex
\usepackage{amsthm}

\usepackage[english]{babel}
\usepackage[utf8]{inputenc} 
\usepackage[T1]{fontenc}    
\usepackage{amsmath}
\usepackage{amssymb}
\usepackage{amsfonts}
\usepackage{multirow}
\usepackage{dsfont}
\usepackage{graphicx}
\usepackage{cases}
\usepackage{color}
\usepackage[colorinlistoftodos]{todonotes}

\usepackage{algorithm}
\usepackage{array}
\newcolumntype{C}[1]{>{\centering\let\newline\\\arraybackslash\hspace{0pt}}m{#1}}

\def\shownotes{1} 
 \ifnum\shownotes=1
\newcommand{\authnote}[2]{{[#1: #2]}}
\else 
\newcommand{\authnote}[2]{{}}
\fi

\newtheorem{theorem}{Theorem}[section]

\newtheorem{question}[theorem]{Question}

\numberwithin{equation}{section}

  \newcommand{\gnorm}[1]{{\vert\kern-0.25ex\vert\kern-0.25ex\vert #1 
		\vert\kern-0.25ex\vert\kern-0.25ex\vert}}

%% file: 0_abstract.tex
\begin{abstract}

Mitigating the negative impact of noisy labels has been a perennial issue in supervised learning.
Robust loss functions have emerged as a prevalent solution to this problem. 
In this work, we introduce the \textit{Variation Ratio} as a novel property related to the robustness of loss functions, and propose a new family of robust loss functions, termed \textit{Variation-Bounded Loss} (VBL), which is characterized by a bounded variation ratio.
We provide theoretical analyses of the variation ratio, proving that a smaller variation ratio would lead to better robustness.
Furthermore, we reveal that the variation ratio provides a feasible method
to relax the symmetric condition and offers a more
concise path to achieve the asymmetric condition.
Based on the variation ratio, we reformulate several commonly used loss functions into a variation-bounded form for practical applications. 
Positive experiments on various datasets exhibit the effectiveness and flexibility of our approach. 
\begin{links}
    \link{Code}{https://github.com/cswjl/variation-bounded-loss}
\end{links}

\end{abstract}

%% file: 1_introduction.tex
\section{Introduction}

In recent years, Deep Neural Networks (DNNs) have demonstrated outstanding performance in a wide range of fields \cite{dlsurvy}.
However, achieving strong performance typically relies on large-scale, high-quality annotated datasets. In practice, real-world datasets often contain substantial label noise due to human error or carelessness \cite{lnlsurvey}.
DNNs may suffer a considerable decline in performance when they overfit to noisy labels \cite{understandingDL}.
As a result, ensuring robust generalization in the presence of noisy labels remains a major challenge. Among the many approaches proposed to address this issue, robust loss functions have remained a popular solution, owing to their simplicity and flexibility \cite{symmetric-condition, GCE, NCE, ALF_TPAMI, OGC}.

Cross Entropy (CE) is the standard loss function for classification tasks due to its strong fitting capability; however, it is also prone to overfitting noisy labels. \citet{symmetric-condition-binary, symmetric-condition} proved that if a loss function satisfies the symmetric condition, it can achieve noise-tolerance. A classic example of a symmetric loss function is Mean Absolute Error (MAE).
However, due to the strict symmetric condition, symmetric loss functions are usually difficult to optimize \cite{GCE, NCE, ALF_ICML}. Therefore, a popular way is to balance the fitting ability and robustness of the loss function by taking an intermediate value between the CE loss and the MAE loss, such as GCE \citep{GCE}, SCE \cite{SCE}, and JS \cite{JS}. However, this approach to improving fitting ability comes at the expense of robustness due to the retained CE property. More specifically, the absolute values of the gradients of these loss functions tend to approach infinity at low prediction probabilities.
This character leads them to still pay too much attention to particularly low-confidence samples which are most likely noisy examples \cite{LC}.
As a result, they do not achieve full noise-tolerance and may overfit to a portion of the noisy labels. Beyond the symmetric condition \cite{symmetric-condition}, \citet{ALF_ICML,ALF_TPAMI} proposed Asymmetric Loss Functions (ALFs), which focus on the loss term with maximum weight in the optimization.
In this way, the impact of noisy components is mitigated, making the loss function inherently noise-tolerant.
Although they proposed the asymmetric condition, they did not provide a straightforward method for its implementation. To date, no design guidelines have been established to help create simpler and more efficient asymmetric loss functions.

In this work, we introduce a new property of loss functions, called \textit{variation ratio}, which involves the robustness of loss functions. 
Then, we propose a new family of robust loss functions, namely \textit{Variation-Bounded Loss} (VBL), whose variation ratio is bounded. We perform rigorous theoretical analyses of the variation ratio.
Firstly, from the perspective of the symmetric condition, we prove that a bounded variation ratio can achieve a relaxed symmetric condition. Moreover, we show that a smaller variation ratio leads to a tighter excess risk bound across various types of label noise.
Secondly, from the perspective of the asymmetric condition, we build a path from the variation ratio to the asymmetric condition. We prove that if the variation ratio is below a certain threshold determined by the noise rate, the variation-bounded loss becomes asymmetric and, consequently, completely noise-tolerant. This gives us a simpler and more efficient way to achieve the asymmetric condition.
Our comprehensive analyses demonstrate that the variation ratio is critical for both symmetric and asymmetric conditions. 
This suggests that the variation ratio can serve as a valuable tool for designing more effective and robust loss functions. Furthermore, we modify several commonly used loss functions to a variation-bounded form for practical applications. The main contributions of our work are highlighted as follows:
\begin{itemize}
\item 
We introduce a novel property related to the robustness of loss functions, namely \textit{variation ratio}, and propose a new family of robust loss functions, termed \textit{Variation-Bounded Loss} (VBL), which have a bounded variation ratio.
\item
We provide comprehensive theoretical analyses of our variation-bounded loss, demonstrating that a small variation ratio is essential for achieving noise-tolerant learning.

\item  
The concise variation ratio can serve as a valuable tool for designing more effective robust loss functions. We develop a series of practical
variation-bounded losses. The results of extensive experiments underscore the superiority of our method.
\end{itemize}

%% file: 2_preliminary.tex
\section{Preliminary}
\paragraph{Supervised Classification.}

For supervised classification tasks, we have a labeled dataset $\gS = {(\rvx_n, y_n)}_{n=1}^N$ for training models. Each pair $(\rvx_n, y_n)$ is i.i.d. drawn from a joint distribution $\gD$ on $\gX \times \gY$, where $\gX \subset \R^d$ is the sample space, $\gY = [K] = \{1, 2, ..., K\}$ is the label space, and $K$ is the number of classes. The classifier $f$ is a model  with a softmax layer, mapping inputs from the sample space to the probability simplex; thus, the predicted label is given by $\hat{y} = \argmax_k f(\rvx)_k$.
We consider a loss function $L: \gU \times \gY \rightarrow \sR$, where $\argmin_{\rvu} L(\rvu, \rve_y) = \rve_y$ and $\rve_y$ is the one-hot vector corresponding to class $y$. The loss function $L(\rvu, \rve_y)$ is monotonically decreasing on the prediction probability $u_y$ of class $y$. For brevity, we abbreviate $L(\rvu, \rve_k)$ as $L(\rvu, k)$ in this paper.
Given a loss function $L \in \gL$ and a classifier $f \in \gF$, the expected risk \cite{bartlett2006risk_bound} is defined as $\gR_L(f) = \mathbb{E}_{(\rvx, y) \sim \gD}[L(f(\rvx), y)]$. The objective of supervised learning is to find a classifier $f^* \in \argmin_{f \in \gF} \gR_L(f)$ that minimizes the expected risk.

\paragraph{Learning with Noisy Labels.}
In the learning with noisy labels scenario, the available training set $\tilde{\mathcal S}=\{(\mathbf x_n,\tilde{y}_n)\}_{n=1}^N$ is noisy rather than a clean set $\mathcal S$. 
For a sample $\rvx$, its true label $y$ will be corrupted into a noisy label $\tilde{y}$ with a conditional probability $\eta_{\rvx, \tilde{y}}=p(\tilde{y}|\rvx, y)$ \cite{LNL}. 
We define the noise rate for $\rvx$ as $\eta_\rvx = \sum_{k \neq y}\eta_{\rvx,k}$.
We mainly consider the following three common types of label noise \citep{IDN-PDN, IDN-beyond, ANL}:
\begin{itemize}
\item{Instance-Dependent Noise:} 
$\eta_{\rvx,y}=1-\eta_\rvx$ and $\sum_{k \neq y}\eta_{\rvx,k}=\eta_\rvx$, where noise rate $\eta_\rvx$ depends on the instance $\rvx$.

\item{Symmetric Noise}:
$\eta_{\rvx,y}=1-\eta$ and  $\eta_{\rvx,k \neq y}=\frac{\eta}{K-1}$, where noise rate $\eta$ is a constant.

\item{Asymmetric Noise:} 
$\eta_{\rvx,y}=1-\eta_y$ and    $\sum_{k \neq y}\eta_{\rvx,k}=\eta_y$, where  noise rate $\eta_y$   depends on the class $y$.

\end{itemize}

For this context, we only have the noisy dataset. Therefore, we focus on minimizing the noisy expected risk as follows:
\begin{equation}
\label{noisy-L-risk} 
\gR_L^\eta(f)=\E_{\gD}[(1-\eta_{\rvx})L(f(\rvx), y)+\sum_{k\neq y}\eta_{\rvx,k} L(f(\rvx),k)],
\end{equation}
where $\sum_{k\neq y}\eta_{\rvx,k} L(f(\rvx),k)$ is the noisy portion that usually damages the performance of DNNs. A loss function $L$ is defined to be \textit{noise-tolerant} \citep{noise-tolerance} if the noise minimizer $f^*_{\eta}\in\argmin_{f \in \gF} \gR_L^\eta(f)$ also minimizes the clean expected risk, i.e., $R_L(f_\eta) = R_L(f)$.



%% file: 3_methodology.tex
\section{Variation-Bounded Loss}

In this section, we provide comprehensive descriptions and theoretical analyses of our variation ratio and variation-bounded loss (VBL), demonstrating that our method achieves robust and efficient learning in a concise manner. 
Detailed proofs are included in the Appendix.

\subsection{Definitions}

First, we introduce the proposed variation ratio and  variation-bounded loss.

\begin{Definition}[Variation Ratio]
For a loss function $L(\rvu, y) = \ell(u_y)$,
we define the variation ratio $v(L)$  as
\begin{equation}
    v(L) = \frac {\max_{u\in(0,1)}|\nabla \ell (u)|}{\min_{u\in(0,1)}|\nabla \ell (u) |},
\end{equation}
where $|\cdot|$ denotes the absolute value and $\nabla \ell = \frac{\partial \ell(u)}{\partial u}$  is the gradient of $\ell$ w.r.t. $u$.
\end{Definition}

\begin{Definition}[Variation-Bounded Loss]
If the variation ratio $v(L) < \infty$, the loss function $L$ is variation-bounded. Conversely, if $v(L) = \infty$, the loss function $L$ is variation-unbounded.
\end{Definition}

A bounded variation ratio ensures that the loss function does not descend excessively within an interval. 
We provide some examples of common loss functions to enhance understanding.
The variation ratio $v(L)$ of Mean Absolute Error (MAE), $L_\text{MAE}(\rvu, y) =2(1 - u_y)$, is 1, indicating that MAE is a variation-bounded loss. Another variation-bounded example is the Exponential Loss (EL), $L_\text{EL}(\rvu, y) = e^{-u_y}$, which has a variation ratio of $e$. In contrast, the variation ratio $v(L)$ of the Cross-Entropy (CE), $L_\text{CE}(\rvu, y) =-\log u_y$, is infinite, indicating that CE is variation-unbounded.

In the following, we will explore the detailed properties of the variation-bounded loss.

\subsection{Symmetric Condition}
Previous works \citep{symmetric-condition-binary, symmetric-condition} theoretically proved that 
symmetric loss functions are inherently robust to label noise under some mild conditions.

\begin{Definition}[Symmetric Condition]
\label{defin: symmetric condition}
A loss function is symmetric if it satisfies
\begin{equation}
    \sum_{k=1}^K L(\rvu, k) = C
\end{equation}
where $C$ is a constant and $k \in [K]$ is the label corresponding to each class.
\end{Definition}

Because the symmetric condition in Definition~\ref{defin: symmetric condition} is overly strict, symmetric losses are challenging to optimize \citep{GCE, ALF_ICML}. A common method to address this issue is to interpolate between the symmetric MAE and the fast-converging CE \cite{GCE,SCE,JS}. Although these robust loss functions can mitigate label noise, we observe that, as they are derived through interpolation with CE, they inevitably inherit a certain property of CE. Specifically, when the predicted probability approaches 0, the absolute value of their gradient approaches $\infty$. Unfortunately, since noisy samples often have low confidence, this characteristic can result in overfitting to some noisy labels \cite{LC}. 

In this work, we propose a novel method to 
relax the symmetric condition instead of interpolating between the MAE and CE.
This method avoids the inherent drawback of CE, which tends to overfit to label noise.
Some studies \citep{symmetric-condition, dividemix, LC} have indicated that bounded losses are more robust than unbounded losses. The bounded property of the loss, \(C_L \leq L(\mathbf{u}, k) \leq C_U\), ensures that \(|\sum_{k=1}^K L(\mathbf{u}, k) - \sum_{k=1}^K L(\mathbf{v}, k)| \leq K (C_U - C_L)\), where $\rvu$ and $\rvv$ are arbitrary vectors in the domain. This property brings the bounded loss closer to meeting the symmetric condition and enhances its robustness compared to unbounded losses such as CE. 
Here, we derive a more efficient bounded property based on the variation ratio.

\begin{Lemma}
\label{lemma: symmetric bound}
For a loss function $L(\rvu, y) = c \cdot \ell(u_y)$,   we have 
\begin{equation}
     \left|\sum_{k=1}^K L(\rvu, k) - \sum_{k=1}^K L(\rvv, k)\right| \le v(L) - 1.
\end{equation}
where $c = \frac{1}{\min_u|\nabla \ell (u)|}$ is a  normalization constant.
\end{Lemma}
$c$ in Lemma~\ref{lemma: symmetric bound} is used to normalize the minimum absolute value of the  gradient to $1$, so that different loss functions can be compared on the same scale.
Lemma~\ref{lemma: symmetric bound} shows that the variation ratio constitutes a sufficient condition for the bounded property and a smaller $v(L)$ must result in better symmetry. 
Specifically, when $v(L)$ reaches its minimum value of 1, the loss is symmetric. And this loss is essentially a linear function, representing a scaled MAE. 

Based on Lemma~\ref{lemma: symmetric bound}, we derive excess risk bounds \citep{bartlett2006risk_bound} under various types of label noise. First, we prove the situation of symmetric noise.
\begin{Theorem}[Excess Risk Bound under Symmetric Noise]
\label{Theorem: symmetric bound}
In a multi-class classification problem, if the loss function $L$ satisfies  $|\sum_{k=1}^K L(\rvu, k) - \sum_{k=1}^K L(\rvv, k)| \le v(L) - 1$, then for symmetric noise satisfying $\eta<1-\frac{1}{K}$, the excess risk bound for $f$ can be expressed as
\begin{equation}
     \gR_L(f^*_\eta)-\gR_L(f^*)\le c(v(L) - 1), 
\end{equation}
where $c=\frac{\eta}{(1-\eta)K-1}$ is a constant, $f^*_\eta$ and $f^*$ denote the global minimum of $\gR_L^\eta(f)$ and $\gR_L(f)$, respectively.
\end{Theorem}

Next, we address the more complex situations of asymmetric and instance-dependent noise.

\begin{Theorem}[Excess Risk Bound under  Asymmetric and Instance-Dependent Noise]
\label{Theorem: asymmetric and idn bound}
In a multi-class classification problem, if the loss function $L$ satisfies $|\sum_{k=1}^K L(\rvu, k) - \sum_{k=1}^K L(\rvv, k)| \le v(L) - 1$,  then for label noise $1-\eta_\rvx > \max_{k \neq y} \eta_{\rvx, k}$,  $\forall \rvx$, if $\gR_L(f^*)$  is minimum, the excess risk bound for $f$ can be expressed as
\begin{equation}
     \mathcal{R}_L(f_\eta^*) -  \mathcal{R}_L(f^*) \le (1 + \frac{c}{a})(v(L) - 1),
\end{equation}
where $c = \mathbb{E}_\mathcal D\left(1-\eta_\rvx\right)$ and $a=\min_{\rvx,k}(1-\eta_\rvx-\eta_{\rvx,k})$ are constants, $f^*_\eta$ and $f^*$ denote the global minimum of $\gR_L^\eta(f)$ and $\gR_L(f)$, respectively.
For asymmetric noise, $\eta_\rvx = \eta_y$, and for instance-dependent noise, $\eta_\rvx = \eta_\rvx$.
\end{Theorem}

Theorem \ref{Theorem: symmetric bound} and \ref{Theorem: asymmetric and idn bound} demonstrate that a smaller variation ratio $v(L)$ results in more robust to label noise. Additionally, the excess risk bound can be controlled by the $v(L)$. 

\subsection{Asymmetric Condition}
Previous works \citep{ALF_ICML, ALF_TPAMI} proposed asymmetric loss functions, which are noise-tolerant to label noise.
However, the proposed asymmetric loss functions are too complex with many hyperparameters, and easily produce the underfitting problem. 
In this subsection, we revisit the asymmetric condition through the variation ratio.
\begin{Definition}[Asymmetric Condition]
On the given weights $w_1, \dots, w_k \geq 0$, where $\exists t \in [K]$, s.t., $w_t > \max_{k \neq t} w_k$, a loss function $L(\rvu, k)$ is called asymmetric if $L$ satisfies
\begin{equation}
\arg\min_{\rvu} \sum_{k=1}^K w_k L(\rvu,k) = \arg\min_{\rvu} L(\rvu,t),
\end{equation}
where we always have $\arg\min_{\rvu} L(\rvu,t) = \rve_t$.
\end{Definition}

Asymmetric loss functions are noise-tolerant under clean-label-dominant noise, i.e., $1-\eta_\rvx > \max_{k \neq y} \eta_{\rvx, k}$,  $\forall \rvx$ \cite{ALF_ICML}. 
Here, we revisit the way to achieve the asymmetric condition and prove that when the variation ratio $v(L)$ is less than a specific constant related to the label distribution, the variation-bounded loss is asymmetric.
\begin{Theorem}
\label{theorem: asymmetric condition}
On the given weights $w_1, \dots, w_k \geq 0$, where $\exists t \in [K]$ and $w_t > \max_{i \neq t} w_i$, a loss function $L(\rvu, k) = \ell(u_k)$ is asymmetric if (1) $\frac{\partial^2 \ell(u)}{\partial u^2} \le 0$ or (2) $v(L) \le \frac{w_t}{w_i}$ for any $i \neq t$.
\end{Theorem}
Condition (1) in Theorem \ref{theorem: asymmetric condition} is not favorable for optimization. Specifically, when $\frac{\partial^2 \ell (u)}{\partial u^2} > 0$, we have a convex loss function, such as CE, which is generally favorable for optimization. When $\frac{\partial^2 \ell (u)}{\partial u^2} = 0$, we have a linear loss function, such as MAE. When $\frac{\partial^2 \ell (u)}{\partial u^2} < 0$, we have a concave loss function, the loss function would be even harder to optimize than linear MAE. 
Therefore, in practice, concave loss functions are generally not considered. Instead, we primarily focus on condition (2) in Theorem \ref{theorem: asymmetric condition}.

For a variation-bounded loss described in Theorem \ref{theorem: asymmetric condition}, if it satisfies $v(L) \le \frac{1-\eta_\rvx}{\max_{k \neq y} \eta_{\rvx, k}}$, i.e., condition (2) in Theorem \ref{theorem: asymmetric condition} for the context of learning with noisy labels,
then the loss function is asymmetric.
For instance, about a 10-classes dataset with 0.8 symmetric noise, if $v(L) \le \frac{1-\eta_\rvx}{\max_{k \neq y} \eta_{\rvx, k}} = \frac{0.2}{0.8/9} \approx 2.25$, then the loss function is asymmetric and therefore noise-tolerant. 
Notably, this constitutes a more relaxed condition compared to the symmetric condition, because it only requires that  $v(L) \le 2.25$, whereas symmetric MAE requires $v(L)$ to equal the minimum value of 1. Thus, variation-bounded losses have better fitting ability than symmetric losses, enabling them to achieve both complete robust and efficient learning simultaneously.

\input{figs/acc_line}

\subsection{Variation-Bounded Loss}
In this subsection, we concisely generalize several commonly used loss functions to a variation-bounded form.  We use $\rvu = f(\rvx)$  to denote the predicted probability after the softmax layer, and $u_y$ is the predicted probability for the label.

\paragraph{Variation Cross Entropy (VCE):}
\begin{equation}
    L_\text{VCE} = - \log(u_y + a),
\end{equation}
where $a \ge 0$ is a hyperparameter. VCE is modified from the CE loss. 
The gradient of VCE is $-\frac{1}{u_y+a}$.
If $a>0$, The variation ratio $v(L_\text{VCE}) = \frac{1+a}{a}$. If $a=0$, the variation ratio $v(L_\text{VCE}) = \infty$, which recovers the CE loss.

\paragraph{Variation Exponential Loss (VEL):}
\begin{equation}
    L_\text{VEL} = a^{-u_y},
\end{equation}
where $a > 1$ is a hyperparameter. VEL is modified from the Exponential Loss (EL). 
The gradient of VEL is $-a^{-u_y}\log a$. 
The variation ratio $v(L_\text{VEL}) = a$, and if $a = e$, it recovers to the Exponential Loss (EL).


\paragraph{Variation Square Log (VSL):}
\begin{equation}
    L_\text{VSL} = [\log(a\cdot u_y+1)-\log2]^2/a,
\end{equation}
where $0 < a \le 1$ is a hyperparameter. VSL is modified from the Square Log Loss. 
The gradient of VSL is $2[\log (a\cdot u_y +1)-\log 2] \cdot \frac{1}{a\cdot u_y +1}$.
If $0 < a < 1$, the variation ratio $v(L_\text{VSL}) = \frac{(a+1)\cdot\log2}{\log2-\log(a+1)}$. If $a=1$, the variation ratio $v(L_\text{VSL}) = \infty$, which recovers the Square Log (SL).

\subsection{More Analyses}
\paragraph{Loss Function Visualization.}
To better analyze the properties of the variation-bounded loss, we visualize the absolute values of gradients and perform experiments on CIFAR-10 with 0.8 symmetric noise, as shown in Figure~\ref{fig:acc_line}. 
Two common scenarios of variation-unbounded losses are observed. The first scenario, exemplified by CE, is shown in Figure~\ref{fig: vce_grad}. As depicted, its gradient approaches 1 for high-confidence samples and approaches $\infty$ for low-confidence samples. The second scenario, represented by the Square Log (SL), is shown in Figure~\ref{fig: vsl_grad}. Here, the gradient decreases to 0 for high-confidence samples and approaches $2\log 2$ for low-confidence samples. In both cases, the variation ratio becomes $\infty$. Intuitively, during optimization, the gradient contribution of low-confidence (noisy) samples becomes disproportionately large, while the contribution of high-confidence (clean) samples becomes too small. Consequently, as training progresses, loss functions like CE and SL overfit to some noisy labels, leading to a decrease in test accuracies, as shown in Figures~\ref{fig: vce_acc} and \ref{fig: vsl_acc}.
In contrast, our variation-bounded losses constrain the variation ratio, yielding a more balanced gradient trade-off between low and high-confidence samples. 
With a simple modification, VCE, VEL, and VSL achieve significantly greater robustness compared to vanilla CE, EL, and SL. 

\paragraph{Hyperparameter Analysis.}
For variation-bounded losses, as can be seen from test accuracies (Figure~\ref{fig: vce_acc},~\ref{fig: vel_acc} and~\ref{fig: vsl_acc}),
a smaller variation ratio ($a\uparrow$ for VCE;  $a\downarrow$ for VEL and VSL) can enhance robustness and achieve noise-tolerant learning. However, a too small variation ratio may reduce the fitting ability. Therefore, it is suggested to choose a moderate variation ratio to achieve both robust and efficient learning. 

\paragraph{Feature Visualization.}
We further compare the robustness of variation-bounded losses and vanilla CE in learning representations. 
We train models on CIFAR-10 with 0.4 symmetric noise and extract the learned features from the test set using t-SNE \citep{tsne}, as shown in Figure~\ref{fig:tsne}. 
For the hyperparameter, we utilize VCE ($a=5$), VEL ($a=1.5$), and VSL ($a=0.1$), refer to the experiment in Figure~\ref{fig:acc_line}.
As can be seen, embeddings generated by CE exhibit evident overfitting to label noise, as evidenced by the blending of embeddings from distinct classes. In contrast, embeddings generated by variation-bounded losses consistently form clear, well-separated clusters. This demonstrates their superior capability to learn robust and distinct representations under label noise.

\input{figs/tsne}

\paragraph{Combination of NCE and VBL.}
Recently, the most advanced robust loss functions often combine their proposed methods with Normalized Cross Entropy (NCE) \citep{NCE}. Notable examples include Active Passive Loss (APL) \citep{NCE}, Asymmetric Loss Functions (ALFs) \citep{ALF_ICML}, and Active Negative Loss (ANL) \citep{ANL}. 
The combination of two different robust loss functions can mutually enhance the optimization processes of each other, thus improving the overall fitting ability of the model \cite{NCE}. 
To obtain better performance and ensure a fair comparison with other combined methods, we also combine the proposed VBL with NCE.
We simply formulate the combination of NCE and VBL as follows:

\begin{equation}
    L_\text{NCE+VBL} = \alpha \cdot L_\text{NCE} + \beta \cdot L_\text{VBL}
\end{equation}

Previous works \cite{ALF_ICML, ALF_TPAMI} have proved that the combination of the symmetric loss and the asymmetric loss remains asymmetric. Because NCE is symmetric, and our VBL is asymmetric. Thus, NCE+VBL is still asymmetric and, therefore, noise-tolerant.

%% file: figs/acc_line.tex
\begin{figure}[!t]
    \centering
    \subfigure[VCE]{
    \label{fig: vce_grad}
    \includegraphics[width=1.55in]{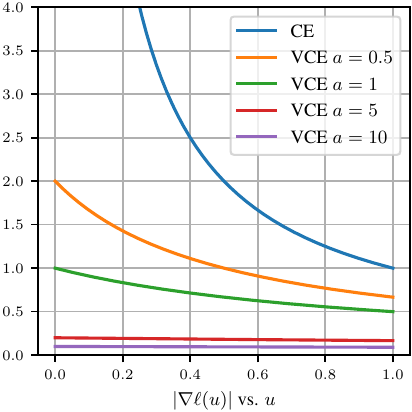}
    }
    \subfigure[VCE]{
    \label{fig: vce_acc}
    \includegraphics[width=1.55in]{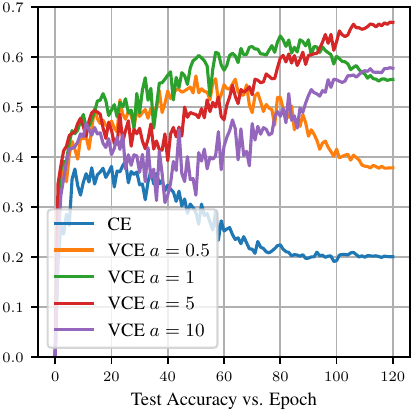}
    } \\
    \subfigure[VEL]{
    \label{fig: vel_grad}
    \includegraphics[width=1.55in]{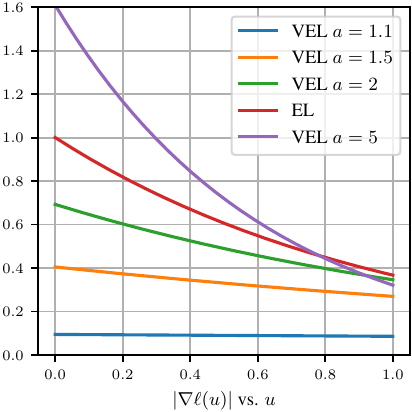}
    }
    \subfigure[VEL]{
    \label{fig: vel_acc}
    \includegraphics[width=1.55in]{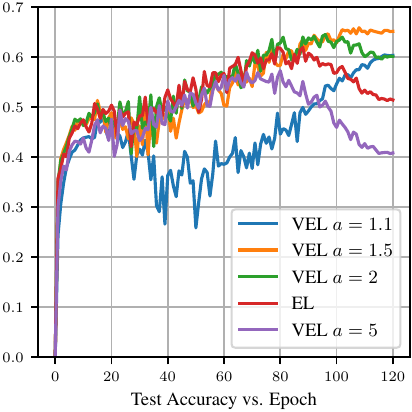}
    } \\
    \subfigure[VSL]{
    \label{fig: vsl_grad}
    \includegraphics[width=1.55in]{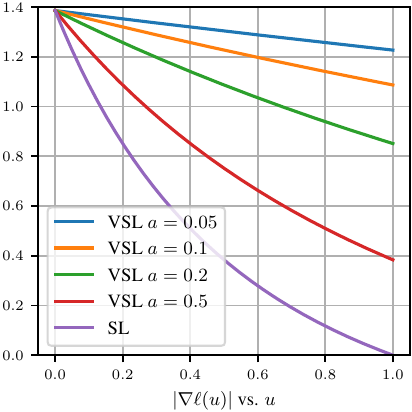}
    }
    \subfigure[VSL]{
    \label{fig: vsl_acc}
    \includegraphics[width=1.55in]{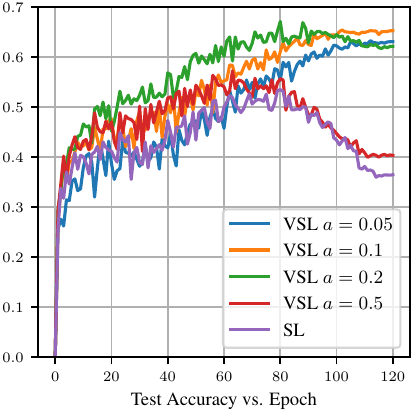}
    }
    \caption{
    \textbf{Left:} Absolute values of gradients, i.e., $|\nabla \ell|$.
    \textbf{Right:} Test accuracies on CIFAR-10 with 0.8 symmetric noise.
    }
    \label{fig:acc_line}
\end{figure}

%% file: figs/tsne.tex
\begin{figure}[!t]
    \centering
    \subfigure[CE]{
    \includegraphics[width=1.55in]{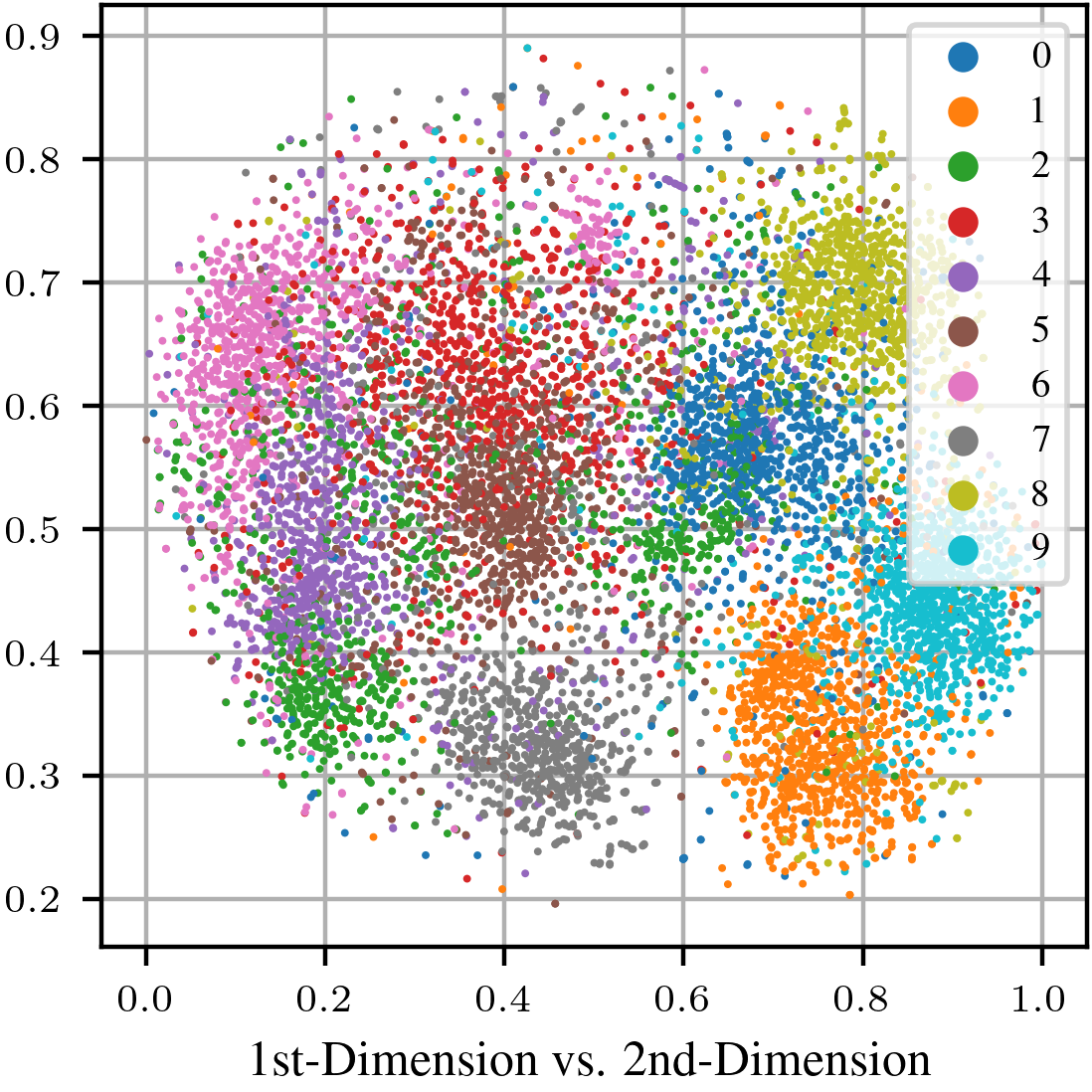}
    }
    \subfigure[VCE]{
    \includegraphics[width=1.55in]{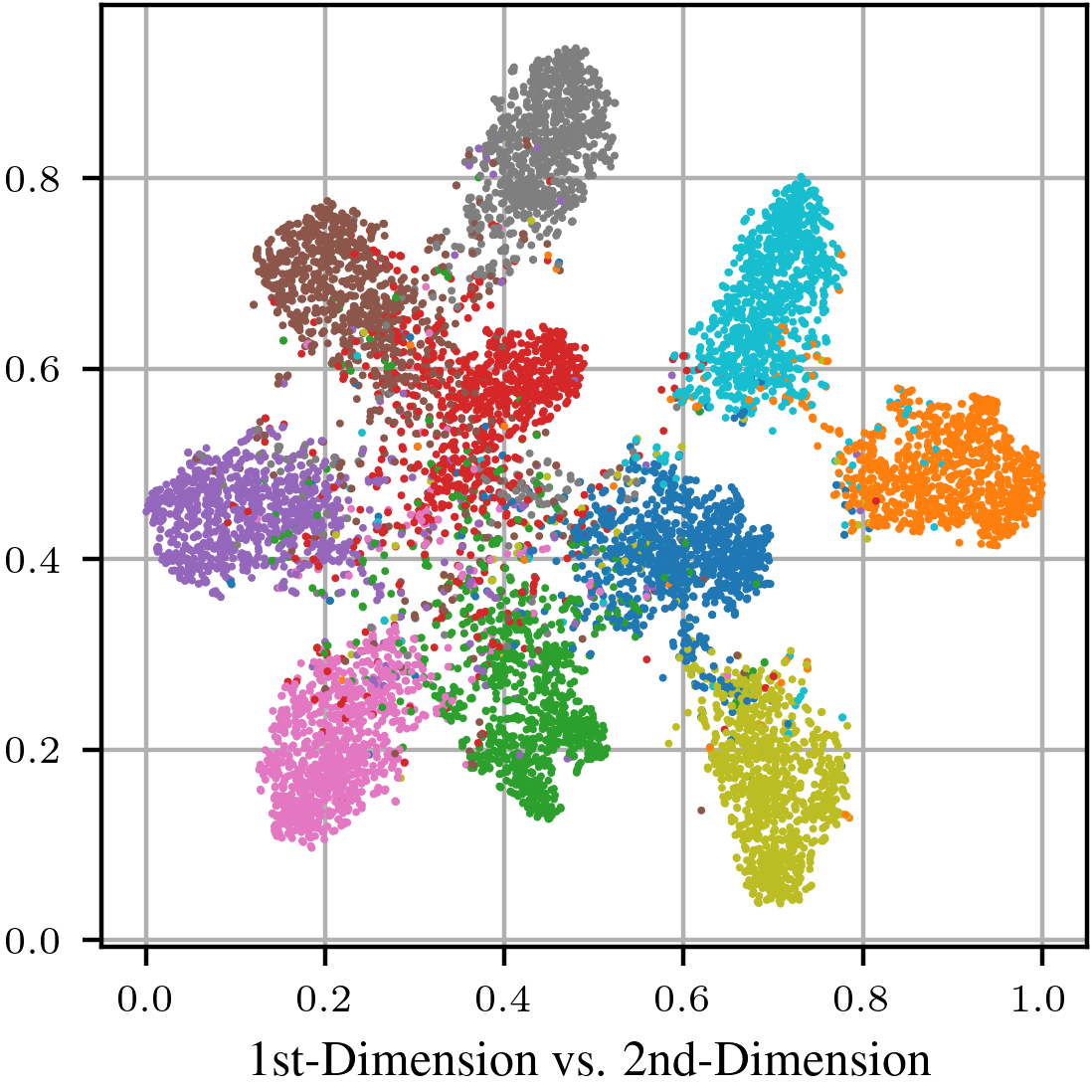}
    } \\
    \subfigure[VEL]{
    \includegraphics[width=1.55in]{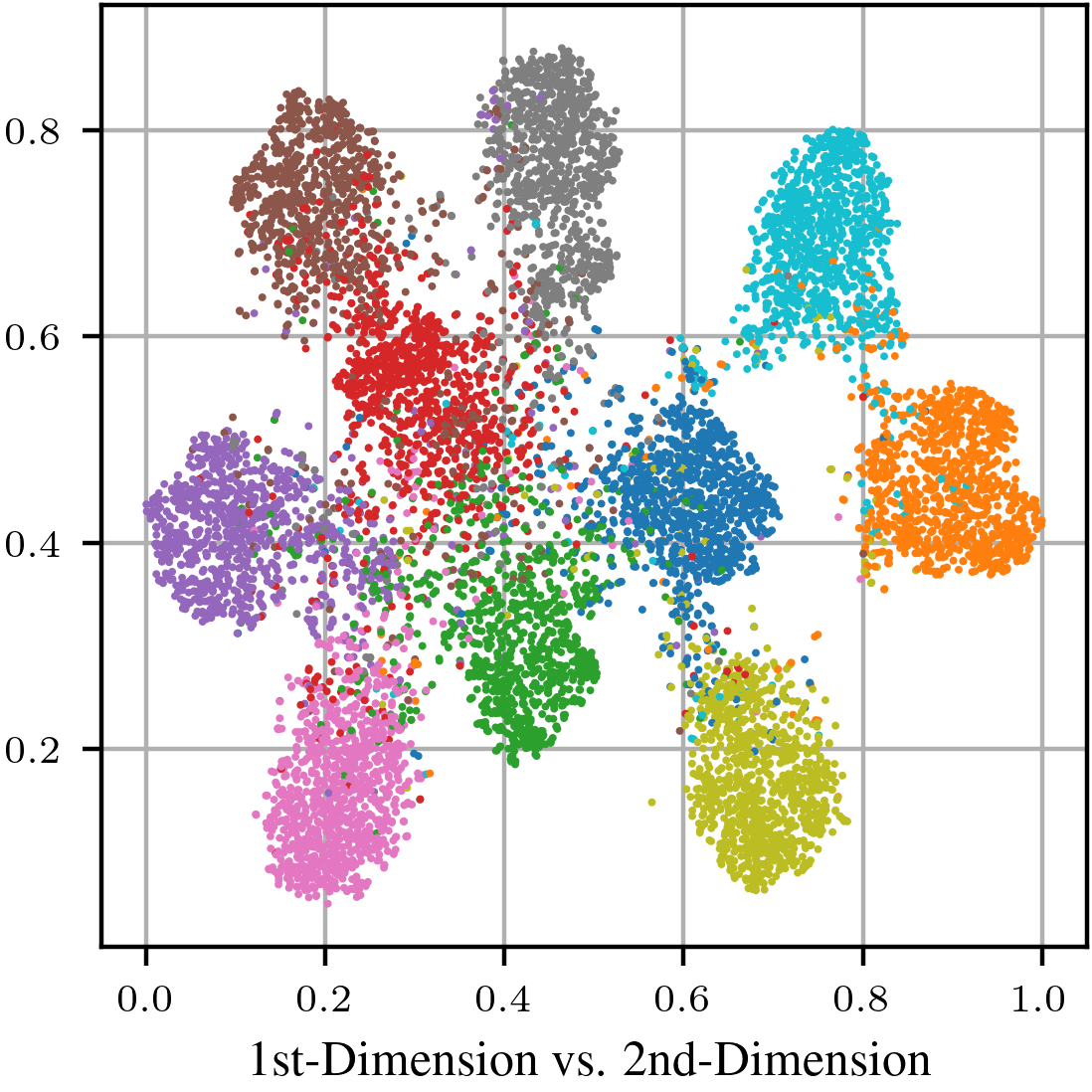}
    }
    \subfigure[VSL]{
    \includegraphics[width=1.55in]{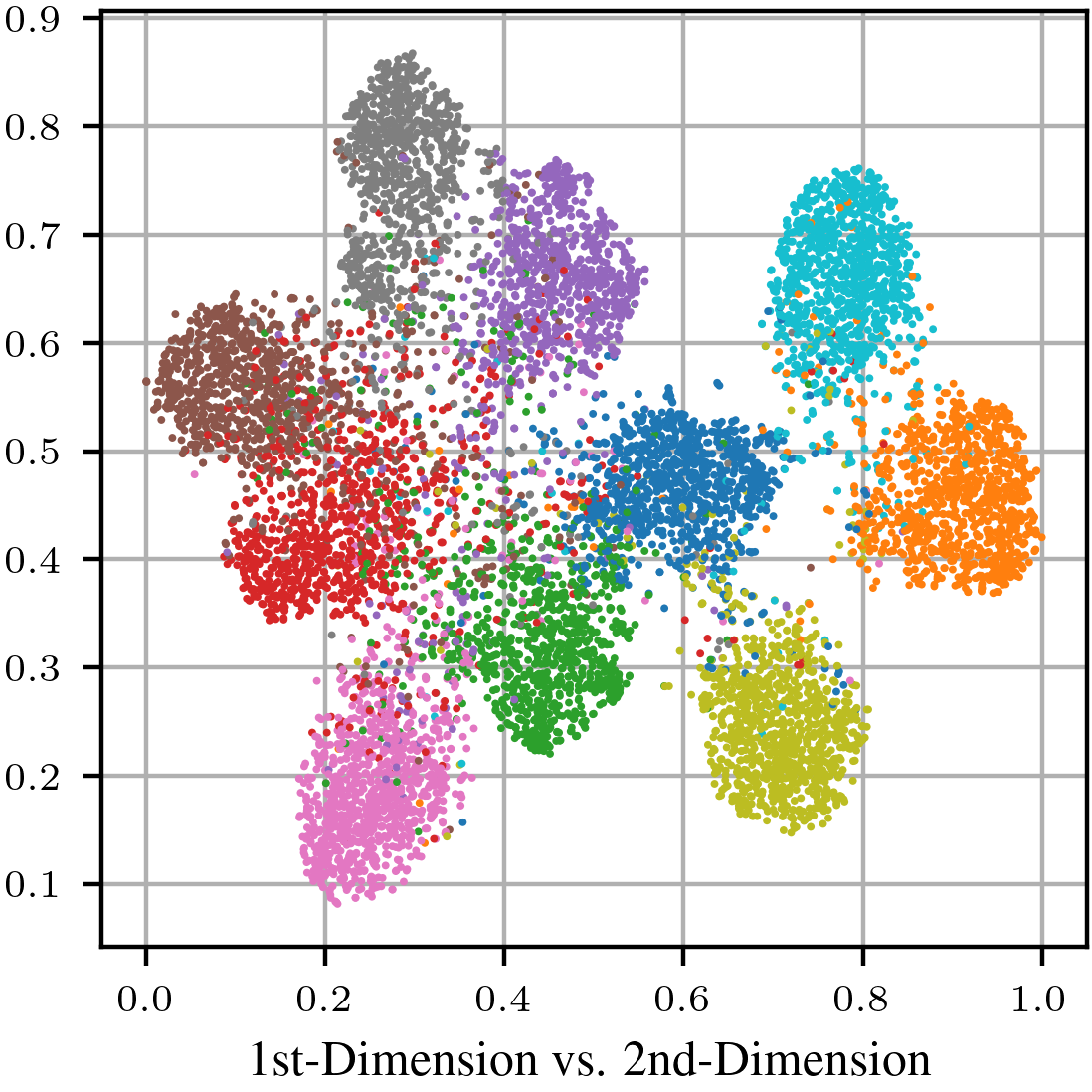}
    }
    \caption{Visualizations of learned features on CIFAR-10 with 0.4 symmetric noise by t-SNE. }
    \label{fig:tsne}
\end{figure}

%% file: 4_experiments.tex
\input{tables/table_idn}

\input{tables/table_S_AS_acc}
\input{tables/table_ablation}

\section{Experiments}
In this section, we provide extensive experiments to evaluate the effectiveness of our variation-bounded loss with various datasets, including benchmark datasets: CIFAR-10 and CIFAR-100 \citep{krizhevsky2009learning}; real-world datasets: WebVision \citep{webvision}, ILSVRC12 \citep{imagenet}, and Clothing1M \citep{clothing1m}. Detailed experiment settings are included in the Appendix.

\subsection{Benchmark Noisy Datasets}

\paragraph{Baselines.}
We experiment with various state-of-the-art methods, including (1)
Cross Entropy (CE);  (2) Generalized Cross Entropy (GCE) \citep{GCE}; (3) Symmetric Cross Entropy (SCE) \citep{SCE}; (4) Active Passive Loss (APL) \citep{NCE}, including NCE+RCE; (5) Asymmetric Loss Functions (ALFs) \citep{ALF_ICML, ALF_TPAMI},  including NCE+AGCE;  (6) LogitClip (LC) \citep{LC}; (7) Active Negative Loss (ANL) \citep{ANL}, including NCE+NNCE; (8) Optimized Gradient Clipping (OGC) \citep{OGC}.  
To avoid confusion and improve performance, we use the NCE+VBL on benchmark datasets.
We follow the same setting as in \citep{NCE, ALF_ICML, ANL}, 
training an 8-layer CNN \cite{cnn} for 120 epochs on CIFAR-10 and a ResNet-34 \citep{resnet} for 200 epochs on CIFAR-100, respectively. 
The experiment results are reported as "mean $\pm$ std" over 3 independent runs.

\input{tables/table_combination}
\input{tables/table_webvision}

\paragraph{Results.}
Table~\ref{tab: idn} and \ref{tab: symm and asymm} showcase the test accuracies of different methods under various types of label noise, including instance-dependent
symmetric, asymmetric, and human-annotated \cite{cifar-n} noise.
Notably, our introduced variation-bounded losses, NCE+VCE, NCE+VEL, and NCE+VSL, exhibit exceptional performance, consistently ranking among the top-3 in most noise types.
In particular, under most difficult 0.6 instance-dependent, 0.8 symmetric, 0.4 asymmetric, and human-annotated noise, our method improves accuracy by 1\%\textasciitilde6\% over previous state-of-the-art methods.
Furthermore, for clean labels, our variation-bounded losses consistently demonstrate superior fitting ability. For instance, on CIFAR-100 clean labels, variation-bounded losses achieve accuracies of 72\%\textasciitilde73\%, whereas other loss functions reach accuracies around 70\%.
These results show that our method achieves superior performance compared to latest advance methods.




\paragraph{Ablation Experiments.}
To further investigate the effect of using variation-bounded loss alone on  benchmark datasets, we conducted additional experiments on symmetric noise, as shown in Table \ref{tab: vce_ablation}.  As can be seen, a simple modification yields a significant improvement in VCE over vanilla CE. In some cases, VCE outperforms the combination of NCE + VCE. Overall, NCE + VCE exhibit a better performance.

\paragraph{Combination of VBL and Other Methods.}

Because of its simplicity and convenience, our variation-bounded loss can be easily integrated with other methods by replacing the original CE loss, without any additional overhead.
We combine our VCE loss with two other approaches: (1) DivideMix \citep{dividemix}, a classic method based on semi-supervised learning, and (2) Negative-LS \citep{Negative-LS}, a label smoothing technique for mitigating label noise.
We conduct experiments on the CIFAR-N datasets, following the same settings as in the original paper \citep{cifar-n},  as shown in Table~\ref{tab: combination}.
As can be seen, our method consistently improves the performance of DivideMix across all scenarios, and yields significant gains for Negative-LS in the most challenging CIFAR-10 worst and CIFAR-100 noisy cases.
These positive results demonstrate that our variation-bounded loss can readily enhance a variety of existing methods, highlighting the broad applicability of our approach.

\subsection{Real-World Noisy Datasets}
In this subsection, we conduct extensive experiments on massive real-world datasets, including WebVision \citep{webvision}, ILSVRC12 (ImageNet) \citep{imagenet}, and Clothing1M \citep{clothing1m}, following the same setting as in \citep{NCE, ANL}. 
For WebVision, we adopt the "Mini" setting from \cite{mentornet}, which utilizes the first 50 classes of the Google subset. We train a ResNet-50 model \cite{resnet} and evaluate it on the same 50 classes on both the ILSVRC12 and WebVision validation sets.
For Clothing1M, we use a ResNet-50 pre-trained on ImageNet. We train it on the noisy dataset of 1 million samples and evaluate it on the clean test set.
\paragraph{Results.}
Table~\ref{tab: webvision} reports the performances on WebVision, ILSVRC12, and 
Clothing1M. As shown, on WebVision and ILSVRC12, our variation-bounded losses, VCE and NCE+VCE, achieve the top-2 best  accuracies, surpassing  previous advanced methods such as NCE+RCE, NCE+AGCE, and NCE+NNCE.  On Clothing1M, the standalone VCE achieves comparable performance to the combined method NCE+NNCE, while our combined method NCE+VCE attains the highest accuracy. These results demonstrate the effectiveness of our approach in real-world scenarios.

%% file: tables/table_idn.tex
\begin{table}[!]
\small
\fontsize{9.2pt}{11.5pt}\selectfont
\centering
\caption{Last epoch test accuracies on instance-dependent noise. 
Top-3 best results are highlighted in \textbf{bold}.
}
\label{tab: idn}
\begin{tabular}{cccc}
\toprule
\multirow{2}{*}{CIFAR-10}  & \multicolumn{3}{c}{Instance-Dependent Noise}                   \\
\cmidrule(lr){2-4}
                           & 20\%                 & 40\%                 & 60\%                 \\
                           \midrule
CE                         & 75.22 $\pm$ 0.09          & 57.33 $\pm$ 0.16          & 37.84 $\pm$ 0.32          \\
GCE                        & 86.86 $\pm$ 0.22          & 82.80 $\pm$ 0.20          & 64.84 $\pm$ 1.04          \\
SCE                        & 86.72 $\pm$ 0.14          & 74.44 $\pm$ 0.39          & 51.15 $\pm$ 0.95          \\
NCE+RCE                    & 89.14 $\pm$ 0.15          & 85.08 $\pm$ 0.39          & 71.55 $\pm$ 0.52          \\
NCE+AGCE                   & 88.97 $\pm$ 0.18          & 84.89 $\pm$ 0.23          & 72.75 $\pm$ 0.34          \\
LC                      & 82.61 $\pm$ 0.23          & 67.82 $\pm$ 0.39          & 43.32 $\pm$ 0.99          \\
NCE+NNCE                   & 89.70 $\pm$ 0.21          & 85.76 $\pm$ 0.28          & 70.61 $\pm$ 1.00          \\
OGC                     & 86.71 $\pm$ 0.22          & 83.33 $\pm$ 0.29          & 64.73 $\pm$ 3.48          \\
\textbf{NCE+VCE}           & \textbf{89.77 $\pm$ 0.01} & \textbf{86.85 $\pm$ 0.23} & \textbf{73.95 $\pm$ 0.26} \\
\textbf{NCE+VEL}           & \textbf{89.80 $\pm$ 0.20} & \textbf{86.93 $\pm$ 0.49} & \textbf{73.85 $\pm$ 0.40} \\
\textbf{NCE+VSL}           & \textbf{89.84 $\pm$ 0.21} & \textbf{86.46 $\pm$ 0.36} & \textbf{75.60 $\pm$ 0.03} \\
\toprule
\multirow{2}{*}{CIFAR-100} & \multicolumn{3}{c}{Instance-Dependent Noise}           \\
\cmidrule(lr){2-4}
                           & 20\%                 & 40\%                 & 60\%                 \\
                           \midrule
CE                         & 56.86 $\pm$ 1.08          & 41.66 $\pm$ 0.33          & 24.47 $\pm$ 1.28          \\
GCE                        & 60.93 $\pm$ 0.92          & 56.81 $\pm$ 1.13          & 41.82 $\pm$ 0.62          \\
SCE                        & 55.70 $\pm$ 1.56          & 40.19 $\pm$ 0.43          & 23.04 $\pm$ 0.88          \\
NCE+RCE                    & 64.63 $\pm$ 0.44          & 56.68 $\pm$ 0.22          & 41.64 $\pm$ 0.77          \\
NCE+AGCE                   & 65.51 $\pm$ 0.24          & 58.40 $\pm$ 0.85          & 42.64 $\pm$ 0.11          \\
LC                      & 56.36 $\pm$ 0.25          & 37.68 $\pm$ 0.21          & 19.28 $\pm$ 0.50          \\
NCE+NNCE                   & 66.63 $\pm$ 0.53          & 61.14 $\pm$ 0.61          & 47.42 $\pm$ 0.62          \\
OGC                     & 64.36 $\pm$ 1.74          & 56.46 $\pm$ 0.44          & 40.04 $\pm$ 0.23          \\
\textbf{NCE+VCE}           & \textbf{69.33 $\pm$ 0.24} & \textbf{64.54 $\pm$ 0.46} & \textbf{54.04 $\pm$ 0.58} \\
\textbf{NCE+VEL}           & \textbf{69.99 $\pm$ 0.31} & \textbf{65.31 $\pm$ 0.39} & \textbf{54.87 $\pm$ 0.43} \\
\textbf{NCE+VSL}           & \textbf{69.97 $\pm$ 0.20} & \textbf{65.44 $\pm$ 0.26} & \textbf{54.43 $\pm$ 0.40} \\
\bottomrule
\end{tabular}
\end{table}

%% file: tables/table_S_AS_acc.tex
\begin{table*}[!t]
\centering
\small
\setlength{\tabcolsep}{1.1mm}
\fontsize{9.2pt}{11.5pt}\selectfont
\caption{Last epoch test accuracies on  symmetric, asymmetric, and human  noise.  
Top-3 best results are highlighted in \textbf{bold}.
}
\label{tab: symm and asymm}
\begin{tabular}{ccccccccc}
\toprule
\multirow{2}{*}{CIFAR-10} & Clean & \multicolumn{4}{c}{Symmetric Noise}                                                   & \multicolumn{2}{c}{Asymmetric Noise}      & Human \\
\cmidrule(lr){2-2}\cmidrule(lr){3-6}\cmidrule(lr){7-8}\cmidrule(lr){9-9}
                           &     0\%                  & 20\%                 & 40\%                 & 60\%                 & 80\%                 & 20\%                 & 40\%                 &   Worst                     \\
                           \midrule
CE                         & 90.47 $\pm$ 0.22             & 75.25 $\pm$ 0.43          & 58.51 $\pm$ 0.52          & 39.21 $\pm$ 0.38          & 19.04 $\pm$ 0.23          & 83.04 $\pm$ 0.17          & 73.70 $\pm$ 0.19          & 62.04 $\pm$ 0.64             \\
GCE                        & 89.08 $\pm$ 0.21             & 87.04 $\pm$ 0.33          & 83.68 $\pm$ 0.17          & 76.30 $\pm$ 0.14          & 42.69 $\pm$ 0.24          & 86.71 $\pm$ 0.15          & 69.12 $\pm$ 0.78          & 77.80 $\pm$ 0.38             \\
SCE                        & 91.48 $\pm$ 0.16             & 87.62 $\pm$ 0.26          & 79.37 $\pm$ 0.40          & 61.38 $\pm$ 0.84          & 27.75 $\pm$ 0.55          & 86.01 $\pm$ 0.21          & 74.17 $\pm$ 0.41          & 73.70 $\pm$ 0.06             \\
NCE+RCE                    & 91.20 $\pm$ 0.15             & 89.17 $\pm$ 0.15          & 85.75 $\pm$ 0.24          & 79.93 $\pm$ 0.30          & 54.04 $\pm$ 2.59          & 88.30 $\pm$ 0.13          & 77.78 $\pm$ 0.19          & 80.09 $\pm$ 0.21             \\
NCE+AGCE                   & 91.05 $\pm$ 0.28             & 89.12 $\pm$ 0.24          & 86.19 $\pm$ 0.19          & 80.28 $\pm$ 0.35          & 45.28 $\pm$ 4.09          & 88.62 $\pm$ 0.09          & 78.50 $\pm$ 0.37          & 80.03 $\pm$ 0.47             \\
LC                         & 90.03 $\pm$ 0.15             & 83.47 $\pm$ 0.53          & 70.27 $\pm$ 0.42          & 46.61 $\pm$ 0.90          & 19.88 $\pm$ 0.83          & 83.31 $\pm$ 0.44          & 73.38 $\pm$ 0.32          & 70.07 $\pm$ 0.22             \\
NCE+NNCE                   & \textbf{91.71 $\pm$ 0.32}             & \textbf{90.01 $\pm$ 0.06} & 87.18 $\pm$ 0.42          & 81.05 $\pm$ 0.36          & 61.31 $\pm$ 2.62          & 88.83 $\pm$ 0.29          & 77.97 $\pm$ 0.12          & 80.52 $\pm$ 0.24             \\
OGC                        & 88.86 $\pm$ 0.12             & 87.17 $\pm$ 0.22          & 83.82 $\pm$ 0.19          & 77.37 $\pm$ 0.26          & 48.09 $\pm$ 0.60          & 86.83 $\pm$ 0.05          & 66.41 $\pm$ 3.20          & 76.31 $\pm$ 3.59             \\
\textbf{NCE+VCE}           & \textbf{91.75 $\pm$ 0.26}             & \textbf{90.03 $\pm$ 0.19} & \textbf{87.73 $\pm$ 0.40} & \textbf{82.33 $\pm$ 0.51} & \textbf{64.49 $\pm$ 0.99} & \textbf{89.72 $\pm$ 0.19} & \textbf{79.16 $\pm$ 0.44} & \textbf{81.28 $\pm$ 0.19}    \\
\textbf{NCE+VEL}           & 91.60 $\pm$ 0.27             & \textbf{90.24 $\pm$ 0.43} & \textbf{87.35 $\pm$ 0.09} & \textbf{82.42 $\pm$ 0.19} & \textbf{64.29 $\pm$ 1.81} & \textbf{89.77 $\pm$ 0.28} & \textbf{80.20 $\pm$ 0.39} & \textbf{81.38 $\pm$ 0.26}    \\
\textbf{NCE+VSL}          & \textbf{91.75 $\pm$ 0.27}             & 89.97 $\pm$ 0.28          & \textbf{87.31 $\pm$ 0.04} & \textbf{82.00 $\pm$ 0.38} & \textbf{62.96 $\pm$ 1.06} & \textbf{89.33 $\pm$ 0.45} & \textbf{79.23 $\pm$ 0.22} & \textbf{81.08 $\pm$ 0.40}    \\
\toprule
\multirow{2}{*}{CIFAR-100} & Clean & \multicolumn{4}{c}{Symmetric Noise}                                                   & \multicolumn{2}{c}{Asymmetric Noise}      & Human \\
\cmidrule(lr){2-2}\cmidrule(lr){3-6}\cmidrule(lr){7-8}\cmidrule(lr){9-9}
                           &     0\%                  & 20\%                 & 40\%                 & 60\%                 & 80\%                 & 20\%                 & 40\%                 &   Noisy                     \\
                           \midrule
CE                         & 71.00 $\pm$ 1.21             & 55.73 $\pm$ 0.73          & 38.03 $\pm$ 2.49          & 23.34 $\pm$ 0.95          & 8.02 $\pm$ 0.22           & 58.03 $\pm$ 0.49          & 41.53 $\pm$ 0.50          & 49.25 $\pm$ 0.38             \\
GCE                        & 64.07 $\pm$ 0.83             & 62.02 $\pm$ 1.88          & 58.03 $\pm$ 1.50          & 46.29 $\pm$ 0.43          & 19.76 $\pm$ 0.82          & 58.57 $\pm$ 1.28          & 41.94 $\pm$ 0.72          & 50.13 $\pm$ 0.57             \\
SCE                        & 70.36 $\pm$ 0.48             & 55.47 $\pm$ 1.14          & 40.04 $\pm$ 0.25          & 22.81 $\pm$ 0.47          & 8.00 $\pm$ 0.18           & 57.40 $\pm$ 0.87          & 41.32 $\pm$ 0.64          & 48.51 $\pm$ 0.07             \\
NCE+RCE                    & 68.54 $\pm$ 0.11             & 64.63 $\pm$ 0.70          & 58.32 $\pm$ 0.34          & 46.40 $\pm$ 1.25          & 25.57 $\pm$ 0.28          & 63.06 $\pm$ 0.13          & 42.29 $\pm$ 0.12          & 54.48 $\pm$ 0.56             \\
NCE+AGCE                   & 68.95 $\pm$ 0.27             & 65.32 $\pm$ 0.29          & 59.40 $\pm$ 0.83          & 47.97 $\pm$ 0.45          & 24.96 $\pm$ 0.42          & 64.21 $\pm$ 0.50          & 44.95 $\pm$ 0.36          & 55.73 $\pm$ 0.17             \\
LC                         & 71.04 $\pm$ 0.25             & 57.37 $\pm$ 0.30          & 37.51 $\pm$ 0.66          & 17.39 $\pm$ 0.52          & 6.84 $\pm$ 0.18           & 56.17 $\pm$ 0.42          & 39.40 $\pm$ 0.18          & 48.15 $\pm$ 0.31             \\
NCE+NNCE                   & 70.27 $\pm$ 0.28             & 67.07 $\pm$ 0.42          & 61.74 $\pm$ 0.20          & 51.50 $\pm$ 0.88          & 28.09 $\pm$ 0.60          & 66.01 $\pm$ 0.25          & 45.92 $\pm$ 0.26          & 56.39 $\pm$ 0.11             \\
OGC                        & 67.99 $\pm$ 1.04             & 63.41 $\pm$ 1.96          & 57.24 $\pm$ 0.60          & 44.71 $\pm$ 3.10          & 14.55 $\pm$ 0.73          & 62.90 $\pm$ 1.11          & 37.56 $\pm$ 0.64          & 53.28 $\pm$ 0.57             \\
\textbf{NCE+VCE}           & \textbf{72.48 $\pm$ 0.43}             & \textbf{69.32 $\pm$ 0.30} & \textbf{64.79 $\pm$ 0.46} & \textbf{57.55 $\pm$ 0.37} & \textbf{30.19 $\pm$ 0.30} & \textbf{68.94 $\pm$ 0.19} & \textbf{51.19 $\pm$ 0.33} & \textbf{57.44 $\pm$ 0.60}    \\
\textbf{NCE+VEL}           & \textbf{73.17 $\pm$ 0.33}             & \textbf{69.97 $\pm$ 0.28} & \textbf{65.43 $\pm$ 0.49} & \textbf{58.09 $\pm$ 0.54} & \textbf{31.41 $\pm$ 0.95} & \textbf{68.99 $\pm$ 0.24} & \textbf{48.27 $\pm$ 0.40} & \textbf{58.31 $\pm$ 0.17}    \\
\textbf{NCE+VSL}           & \textbf{72.48 $\pm$ 0.15}             & \textbf{70.30 $\pm$ 0.29} & \textbf{65.54 $\pm$ 0.10} & \textbf{57.57 $\pm$ 0.26} & \textbf{31.37 $\pm$ 1.20} & \textbf{69.17 $\pm$ 0.03} & \textbf{46.58 $\pm$ 0.41} & \textbf{58.62 $\pm$ 0.20}   \\
\bottomrule
\end{tabular}
\end{table*}

%% file: tables/table_ablation.tex
\begin{table*}[!t]
\small
\setlength{\tabcolsep}{1.1mm}
\fontsize{9.2pt}{11.5pt}\selectfont
\centering
\caption{Last epoch test accuracies of ablation experiment on symmetric noise.  
Best results are highlighted in \textbf{bold}.
}
\label{tab: vce_ablation}
\begin{tabular}{ccccccccc}
\toprule
\multirow{2}{*}{Method} & \multicolumn{4}{c}{CIFAR-10 Symmetric   Noise}                                        & \multicolumn{4}{c}{CIFAR-100 Symmetric   Noise}                                       \\
\cmidrule(lr){2-5}\cmidrule(lr){6-9}
                        & 20\%                 & 40\%                 & 60\%                 & 80\%                 & 20\%                 & 40\%                 & 60\%                 & 80\%                 \\
                        \midrule
CE                      & 75.25 $\pm$ 0.43          & 58.51 $\pm$ 0.52          & 39.21 $\pm$ 0.38          & 19.04 $\pm$ 0.23          & 55.73 $\pm$ 0.73          & 38.03 $\pm$ 2.49          & 23.34 $\pm$ 0.95          & 8.02 $\pm$ 0.22           \\
NCE                     & 73.22 $\pm$ 0.35          & 69.37 $\pm$ 0.22          & 62.47 $\pm$ 0.85          & 41.20 $\pm$ 1.25          & 25.43 $\pm$ 0.91          & 20.26 $\pm$ 0.25          & 14.66 $\pm$ 1.04          & 8.82 $\pm$ 0.47           \\
\textbf{VCE}            & \textbf{90.44 $\pm$ 0.51} & 87.29 $\pm$ 0.22          & 82.28 $\pm$ 0.29          & 63.77 $\pm$ 2.13          & 65.42 $\pm$ 0.81          & 60.39 $\pm$ 0.84          & 49.72 $\pm$ 1.26          & \textbf{33.01 $\pm$ 0.83} \\
\textbf{NCE+VCE}        & 90.03 $\pm$ 0.19          & \textbf{87.73 $\pm$ 0.40} & \textbf{82.33 $\pm$ 0.51} & \textbf{64.49 $\pm$ 0.99} & \textbf{69.32 $\pm$ 0.30} & \textbf{64.79 $\pm$ 0.46} & \textbf{57.55 $\pm$ 0.37} & 30.19 $\pm$ 0.30       \\
\bottomrule
\end{tabular}

\end{table*}

%% file: tables/table_combination.tex
\begin{table*}[!t]

\small
\fontsize{9.2pt}{11.5pt}\selectfont
\centering
\caption{Best epoch test accuracies on CIFAR-N. 
Better results are highlighted in \textbf{bold}.
}
\label{tab: combination}

\begin{tabular}{ccccccc}
\toprule
\multirow{2}{*}{Method} & \multicolumn{5}{c}{CIFAR-10N}                                                                      & CIFAR-100N \\
\cmidrule(lr){2-6}\cmidrule(lr){7-7}
                               & Aggregate           & Random 1            & Random 2            & Random 3            & Worst               & Noisy               \\
                               \midrule
DivideMix                      & 95.01  $\pm$ 0.71          & 95.16  $\pm$ 0.19          & 95.23  $\pm$ 0.07          & 95.21  $\pm$ 0.14          & 92.56  $\pm$ 0.42          & 71.13  $\pm$ 0.48          \\
\textbf{DivideMix+VCE}         & \textbf{95.97  $\pm$ 0.14} & \textbf{96.24  $\pm$ 0.15} & \textbf{96.07  $\pm$ 0.12} & \textbf{96.00  $\pm$ 0.19} & \textbf{93.70  $\pm$ 0.29} & \textbf{71.92  $\pm$ 0.36} \\
\midrule
Negative-LS                    & \textbf{91.97  $\pm$ 0.46} & 90.29  $\pm$ 0.32          & \textbf{90.37  $\pm$ 0.12} & 90.13  $\pm$ 0.19          & 82.99  $\pm$ 0.36          & 58.59  $\pm$ 0.98          \\
\textbf{Negative-LS+VCE}       & 91.72  $\pm$ 0.16          & \textbf{90.75  $\pm$ 0.22} & 90.30  $\pm$ 0.27          & \textbf{90.34  $\pm$ 0.26} & \textbf{84.17  $\pm$ 0.42} & \textbf{61.93  $\pm$ 0.22} \\
\bottomrule
\end{tabular}

\end{table*}

%% file: tables/table_webvision.tex
\begin{table*}[!t]
\small
\fontsize{9.2pt}{11.5pt}\selectfont
\centering
\caption{Last epoch test accuracies on WebVision, ILSVRC12, and Clothing1M. 
Top-2 best results are highlighted in \textbf{bold}.
Baseline results are obtained from \citep{ANL} with the same setting.
}
\label{tab: webvision}
\begin{tabular}{c|ccccccccc}
\toprule
\textbf{Method}       & CE    & GCE   & SCE              & NCE+RCE & NCE+AGCE         & NCE+NNCE       & NFL+NNFL         & \textbf{VCE}            & \textbf{NCE+VCE}        \\
\midrule
\textbf{WebVision}  & 61.2  & 59.44 & 68               & 64.92   & 63.92            & 67.44          & 68.32            & \textbf{69.69 $\pm$ 0.31} & \textbf{69.00 $\pm$ 0.60} \\
\midrule
\textbf{ILSVRC12}   & 58.64 & 56.56 & 62.6             & 62.4    & 60.76            & 65             & 65.56            & \textbf{66.05 $\pm$ 0.25} & \textbf{65.85 $\pm$ 0.20} \\
\midrule
\textbf{Clothing1M} & 68.07 & 68.94 & - & 69.07   & - & 69.93 & - & \textbf{69.96 $\pm$ 0.30}          & \textbf{70.05 $\pm$ 0.36} \\
\bottomrule
\end{tabular}
\end{table*}

%% file: 5_conclusion.tex

\section{Conclusion}
This paper introduces the \textit{Variation Ratio} property of loss functions and proposes a new category of robust loss functions known as \textit{Variation-Bounded Loss} (VBL).
We demonstrate that a smaller variation ratio represents better robustness.
Moreover, we reveal that variation-bounded losses have a better way to relax the symmetric condition and more straightforwardly achieve the asymmetric condition.
Our concise robust loss functions have shown positive results in mitigating label noise across diverse noise types.
We believe that these loss functions can be widely applied in scenarios where it is difficult to obtain precise annotations. 
Additionally, we anticipate that the variation ratio will serve as a valuable tool for designing more effective robust loss functions.


%% file: 6_appendix.tex
\newpage
\appendix
\onecolumn
\section*{\centering\LARGE Appendix
for “Variation-Bounded Loss for Noise-Tolerant Learning”
}
\label{sec: appendix}

\vspace{30pt}

\section{Related Work}

Robust loss functions are a important research area in learning with noisy labels. \citet{symmetric-condition-binary, symmetric-condition} demonstrated that symmetric loss functions, such as Mean Absolute Error (MAE), are inherently noise-tolerant to label noise. However, due to the strict symmetric condition, these loss functions are often challenging to optimize \cite{GCE, NCE, ALF_ICML}.
To address this limitation, many studies have sought to relax the symmetric condition. One widely adopted approach is to interpolate between Cross Entropy (CE) and MAE, resulting in loss functions such as Generalized Cross Entropy (GCE) \citep{GCE}, Symmetric Cross Entropy (SCE) \cite{SCE}, Taylor Cross Entropy (Taylor-CE) \cite{Taylor-CE}, and Jensen-Shannon Divergence Loss (JS) \cite{JS}. Another approach involves relaxing the symmetric condition by approximating the one-hot vector, as seen in methods like Sparse Regularization (SR) \cite{SR} and $\epsilon$-Softmax \cite{wangepsilon}.
Additionally, LogitClip (LC) \cite{LC} and Optimized Gradient Clipping (OGC) \cite{OGC} achieve a relaxed symmetric condition by clamping the logits and gradient, respectively.
Unlike methods that relax the symmetric condition, Active Passive Loss (APL) \cite{NCE} and Active Negative Loss (ANL) \cite{ANL} leverage two distinct symmetric losses simultaneously to enhance the fitting ability.
Beyond the symmetric condition \cite{symmetric-condition}, \citet{ALF_ICML,ALF_TPAMI} proposed Asymmetric Loss Functions (ALFs), which are robust to label noise under a more favorable condition, such as Asymmetric Generalized Cross Entropy (AGCE). Recently, a new asymmetric loss function,  Asymmetric Mean Square Error (AMSE) \cite{AMSE}, has been proposed, extending the asymmetric loss function to a passive loss type.

\section{Visualization of Expected Calibration Error}
To provide a more comprehensive evaluation of our variation-bounded losses, we visualize the Expected Calibration Error (ECE). Similar to Figure 1, we conduct experiments on CIFAR-10 with 0.8 symmetric noise, using VCE (a = 5), VEL (a = 1.5), and VSL (a = 0.1). The results are shown in Figure~\ref{fig:ece}. As illustrated, our method not only achieves higher accuracy (ACC) but also yields a smaller ECE.

\input{figs/ece}

\section{Proofs}
\label{sec: ap proof}
\begin{Lemma}
For a loss function $L(\rvu, y) = c \cdot \ell(u_y)$,   we have 
\begin{equation}
     \left|\sum_{k=1}^K L(\rvu, k) - \sum_{k=1}^K L(\rvv, k)\right| \le v(L) - 1.
\end{equation}
where $c = \frac{1}{\min_u|\nabla \ell (u)|}$ is a  normalization constant.
\end{Lemma}
\begin{proof}
For any $\rvu$, we have $
\sum_{k=1}^K\ell(0) - \sum_{k=1}^K u_k\max_u|\nabla \ell| \le \sum_{k=1}^K \ell(u_k) \le \sum_{k=1}^K\ell(0) - \sum_{k=1}^K u_k\min_u|\nabla \ell|$; this is 
\begin{equation}
K\cdot \ell(0) - \max_u|\nabla \ell| \le \sum_{k=1}^K \ell(u_k) \le K\cdot \ell(0) -\min_u|\nabla \ell|. 
\end{equation}
Hence, we have $ 
 |\sum_{k=1}^K \ell(u_k) - \sum_{k=1}^K\ell(v_k)| \le \max_u|\nabla \ell| - \min_u|\nabla \ell|$ and $|\sum_{k=1}^K L(\rvu, k) - \sum_{k=1}^K L(\rvv, k)| \le v(L) - 1.$
\end{proof}

\begin{Theorem}[Excess Risk Bound under Symmetric Noise]

In a multi-class classification problem, if the loss function $L$ satisfies  $|\sum_{k=1}^K L(\rvu, k) - \sum_{k=1}^K L(\rvv, k)| \le v(L) - 1$, then for symmetric noise satisfying $\eta<1-\frac{1}{K}$, the excess risk bound for $f$ can be expressed as
\begin{equation}
     \gR_L(f^*_\eta)-\gR_L(f^*)\le c(v(L) - 1), 
\end{equation}
where $c=\frac{\eta}{(1-\eta)K-1}$ is a constant, $f^*_\eta$ and $f^*$ denote the global minimum of $\gR_L^\eta(f)$ and $\gR_L(f)$, respectively.

\end{Theorem}
\begin{proof}

For symmetric noise, we have
\begin{equation}
    \begin{aligned}
    R_L^\eta(f^*)&=\mathbb E_{\mathbf x,y}\big[(1-\eta)L(f^*(\mathbf x),y)+\frac{\eta}{K-1}\sum_{k\not = y}L(f^*(\mathbf x), k)\big]\\
    &=(1-\frac{\eta K}{K-1})R_L(f^*) + \frac{\eta}{K-1}\mathbb E_{\mathbf x,y}\left[\sum_{k=1}^K L(f^*(\mathbf x),k) \right]\\
    \end{aligned}
\end{equation}
Similarly, we can obtain
\begin{equation}
    R_L^\eta(f^*_\eta) =(1-\frac{\eta K}{K-1})R_L(f^*_\eta) + \frac{\eta}{K-1}\mathbb E_{\mathbf x,y}\left[\sum_{k=1}^K L(f^*_\eta(\mathbf x),k) \right]
\end{equation}

Since $f^*_\eta=\mathop{\arg\min_u} R_L^\eta(f)$, and $f^*=\mathop{\arg\min_u} R_L(f)$, we have
\begin{equation}
    \begin{aligned}
     &R_L^\eta(f^*_\eta)-R_L^\eta(f^*)\\
     &=(1-\frac{\eta K}{K-1})(R_L(f^*_\eta) - R_L(f^*))+\frac{\eta}{K-1} \mathbb E_{\rvx, y}[\sum_{k=1}^KL(f^*_\eta(\rvx), k) - \sum_{k=1}^KL(f^*(\rvx), k)]
     \le 0\\
     \Rightarrow & R_L(f^*_\eta)-R_L(f^*)\le \frac{\eta}{(1-\eta)K-1}(v(L)-1)
    \end{aligned}
\end{equation}
where we have used the fact that $1-\frac{\eta K}{K-1}>0$.
\end{proof}

\begin{Theorem}[Excess Risk Bound under Asymmetric and Instance-Dependent Noise]
In a multi-class classification problem, if the loss function $L$ satisfies $|\sum_{k=1}^K L(\rvu, k) - \sum_{k=1}^K L(\rvv, k)| \le v(L) - 1$,  then for label noise $1-\eta_\rvx > \max_{k \neq y} \eta_{\rvx, k}$,  $\forall \rvx$, if $\gR_L(f^*)$  is minimum, the excess risk bound for $f$ can be expressed as
\begin{equation}
     \mathcal{R}_L(f_\eta^*) -  \mathcal{R}_L(f^*) \le (1 + \frac{c}{a})(v(L) - 1),
\end{equation}
where $c = \mathbb{E}_\mathcal D\left(1-\eta_\rvx\right)$ and $a=\min_{\rvx,k}(1-\eta_\rvx-\eta_{\rvx,k})$ are constants, $f^*_\eta$ and $f^*$ denote the global minimum of $\gR_L^\eta(f)$ and $\gR_L(f)$, respectively.
For asymmetric noise, $\eta_\rvx = \eta_y$, and for instance-dependent noise, $\eta_\rvx = \eta_\rvx$.

\end{Theorem}
\begin{proof}
   For asymmetric and instance-dependent noise, we have $$
\begin{aligned} 
R_{L}^\eta(f) & =\mathbb{E}_\mathcal D\left[\left(1-\eta_\rvx\right) L(f(\rvx), y)\right]+\mathbb{E}_\gD[\sum_{k \neq y} \eta_{\rvx, k} L(f(\rvx), k)] \\
& = \mathbb{E}_\gD\left[(1-\eta_\rvx)\left(\sum_{k=1}^KL(f(\rvx), y)-\sum_{k \neq y} L(f(\rvx), k)\right)\right]+\mathbb{E}_\gD\left[\sum_{k \neq y} \eta_{\rvx, k} L(f(\rvx), k)\right] \\ 
& =\sum_{k=1}^KL(f(\rvx), y) \mathbb{E}_\gD(1-\eta_\rvx)-\mathbb{E}_\gD\left[\sum_{k \neq y}(1-\eta_\rvx-\eta_{\rvx,k}) L(f(\rvx), k)\right]\\
\end{aligned}
$$

hence,
$$
\begin{aligned}
\left(R_{L}^\eta\left(f^*\right)-R_{L}^\eta(f^*_\eta)\right) = & (\sum_{k=1}^KL(f^*(\rvx), y) - \sum_{k=1}^KL(f^*_\eta(\rvx), y)) \mathbb{E}_\gD(1-\eta_\rvx)+ \\
& \mathbb{E}_\mathcal D \sum_{k \neq y}(1-\eta_\rvx-\eta_{\rvx, k})\left[L(f^*_\eta(\rvx), k)-L\left(f^*(\rvx), k\right)\right]
\end{aligned}
$$
According to the assumption $R_L(f^*)$ is minimum, we have $L(f^*(\rvx), y)$ is minimum then $L(f^*(\rvx), k)$ is 
maximum where $k \neq y$. Since $L(f^*_\eta(\rvx), k)-L(f^*(\rvx), k)\leq 0$ where $k \neq y$, the second term on the right of the inequality is a non-positive value. And $R_{L}^\eta\left(f^*\right)-R_{L}^\eta(f^*_\eta) \geq 0$. So we have

$$
\left|\mathbb{E}_{\mathcal{D}}\sum_{k\neq y} (1-\eta_\rvx-\eta_{\rvx,k})\left(L(f_{\eta}^*(\rvx),k)-L(f^*(\rvx),k)\right)\right|\le c(v(L) - 1),
$$
where $c = \mathbb{E}_\mathcal D\left(1-\eta_\rvx\right)$. 

Let $a=\min_{\rvx,k}(1-\eta_\rvx-\eta_{\rvx,k})$, 
we have $ \left|\mathbb{E}_{\mathcal{D}}\sum_{k\neq y}\left(L(f_{\eta}^*(\rvx),k)-L(f^*(\rvx),k)\right)\right|\le \frac{c(v(L)-1)}{a}$. Note that$|\sum_{k}\left(L(f_{\eta}^*(\rvx),k)-L(f^*(\rvx),k)\right)|\le v(L) - 1$, then we obtain
$$
\left|\mathbb{E}_{\mathcal{D}}\left(L(f_{\eta}^*(\rvx),y)-L(f^*(\rvx),y)\right)\right|\le (v(L)-1) + \frac{c(v(L)-1)}{a},
$$
that is, $ \mathcal{R}_L(f_\eta^*) -  \mathcal{R}_L(f^*) \le (1 + \frac{c}{a})(v(L) - 1)$.
\end{proof}

\begin{Theorem}
    
On the given weights $w_1, \dots, w_k \geq 0$, where $\exists t \in [K]$ and $w_t > \max_{i \neq t} w_i$, a loss function $L(\rvu, k) = \ell(u_k)$ is asymmetric if (1) $\frac{\partial^2 \ell(u)}{\partial u^2} \le 0$ or (2) $v(L) \le \frac{w_t}{w_i}$ for any $i \neq t$.

\end{Theorem}
\begin{proof}
According to \citep{ALF_ICML}, for any \( w_1 > w_2 \geq 0 \), if \( \ell \) satisfies \( w_1 \ell(u_1) + w_2 \ell(u_2) = w_1 \ell(u_1 + u_2) + w_2 \ell(0) \), and the equality holds only if \( u_2 = 0 \), then \( L \) is completely asymmetric.

This is
$$
\begin{aligned}
&w_1(\ell(u_1) - \ell(u_1+u_2)) \ge w_2(\ell(0) - \ell(u_2))\\
\Rightarrow&w_1\frac {(\ell(u_1) - \ell(u_1+u_2))} {u_2} \ge w_2 \frac {(\ell(0) - \ell(u_2))} {u_2}
\end{aligned}
$$

If $\frac{\partial^2 \ell(u_k)}{\partial u_k^2} \le 0$, we have $\ell(u_1) - \ell(u_1+u_2) \ge \ell(0) - \ell(u_2)$, because $\nabla \ell(x+u_1) \le \nabla \ell(x)$, thus established.
In other cases, according to Lagrange's mean value theorem, we  have
\begin{equation}
\label{eq:lagrange}
    w_1 (|\nabla \ell(\xi_1)|) \ge w_2 (|\nabla \ell(\xi_2)|)
\end{equation}
where $\xi_1 \in [u_1,  u_1+u_2]; \xi_2 \in [0, u_2]$. we have
$
\frac {w_1} {w_2} \ge \frac {|\nabla\ell(\xi_2)|} {|\nabla\ell(\xi_1)|}
$, if $\frac {w_1} {w_2} \ge \frac {\max_u|\nabla\ell|} {\min_u|\nabla\ell|} = v(L)$, E.q. \ref{eq:lagrange} is true.

\end{proof}

\section{Experiments}
\label{sec: ap experiment}
\subsection{Benchmark Noisy Datasets}

\paragraph{Noise Generation.}

The noisy labels are generated following standard approaches in previous works  \citep{NCE, ALF_ICML,ANL}. For symmetric noise, we flip the labels in each class randomly to incorrect labels of other classes. For asymmetric noise, we flip the labels within a specific set of classes. For CIFAR-10, flipping TRUCK $\rightarrow$ AUTOMOBILE, BIRD $\rightarrow$ AIRPLANE, DEER $\rightarrow$ HORSE, CAT $\leftrightarrow$ DOG. For CIFAR-100, the 100 classes are grouped into 20 super-classes with each has 5 sub-classes, and each class are flipped within the same super-class into the next in a circular fashion. For instance-dependent noise, we follow the same approach in PDN \citep{IDN-PDN} for generating label noise.

\paragraph{Experiment Setting.}
All experiments are  implemented by PyTorch and are conducted on NVIDIA RTX 4090.
We follow the experiment setting in previous works \citep{NCE, ALF_ICML,ANL}. An 8-layer CNN \cite{cnn} is used for CIFAR-10, and a ResNet-34 \cite{resnet} is used for CIFAR-100. For CIFAR-10 and CIFAR-100, the networks are trained for 120 and 200 epochs with batch size 128. We use SGD optimizer with momentum 0.9 and L1 weight decay  $5\times10^{-5}$ and $5\times10^{-6}$ for CIFAR-10 and CIFAR-100. 
The learning rate is set to 0.01 for CIFAR-10 and 0.1 for CIFAR-100 with cosine annealing. 
Typical data augmentations including random shift and horizontal flip are applied. All the experiments used fixed seeds 123, 124, and 125 for three repeated trials.


\paragraph{Parameters Setting.}

We use the best parameters which match their original papers for all baselines.
Specifically, for GCE, we set $q=0.9$ for CIFAR-10 and $q=0.7$ for CIFAR-100. 
For SCE, we set $A=-4$, and $\alpha=0.1$, $\beta=1$ for CIFAR-10, $\alpha=6$, $\beta=0.1$ for CIFAR-100. 
For NCE+RCE, we set $A=-4$, $\alpha=1,\beta=1$ for CIFAR-10 and $\alpha=10$, $\beta=0.1$ for CIFAR-100. 
For NCE+AGCE, we set $a=6, q=1.5, \alpha=1, \beta=4$ for CIFAR-10 and $a=1.8, q=3, \alpha=10, \beta=0.1$ for CIFAR-100. 
For LC, we use CE+LC. We set $\delta = 2.5$ for CIFAR-10 and $\delta = 0.5$ for CIFAR-100.
For NCE+NNCE, we set $\alpha=5, \beta=5$ for CIFAR-10 and $\alpha=10, \beta=1$ for CIFAR-100. 
For OGC, we use CE+OGC and search the best $\epsilon_0$ in $[1, 5, 10, 20, 50, 100]$ following the original parer. We set $\epsilon_0=1$ for 0.6 and 0.8 noise and $\epsilon_0=5$ for others. 
For NCE+VCE, we set $\alpha=1, \beta=10, a=4$ for CIFAR-10 and $\alpha=5, \beta=1, a=0.4$ for CIFAR-100.
For NCE+VEL, we set $\alpha=1, \beta=10, a=1.2$ for CIFAR-10 and $\alpha=5, \beta=1, a=5$ for CIFAR-100.
For NCE+VSL, we set $\alpha=1, \beta=5, a=0.05$ for CIFAR-10 and $\alpha=5, \beta=1, a=0.65$ for CIFAR-100. Our experiments suggest that more complex datasets benefit from a higher $\alpha$.




\subsection{Real-World Noisy Datasets}
\paragraph{Experiment Setting for WebVision and ILSVRC12.}
We follow the same experiment setting in \citep{ANL}. We use the "Mini" setting described in \citep{mentornet}, which includes the first 50 classes of WebVision. We train a model on Webvision and evaluate the trained model on the same 50 concepts on the corresponding WebVision and ILSVRC12 validation sets. We train a ResNet-50 using SGD for 250 epochs with initial learning rate 0.4, nesterov momentum 0.9 and L1 weight decay $6 \times 10^{-5}$ and batch size 512. The learning rate is multiplied by 0.97 after each epoch of training. All the images are resized to $224 \times 224$. Typical data augmentations including random shift, color jittering, and  horizontal flip are applied.

\paragraph{Parameters Setting for WebVision.}
For GCE, we set $q=0.7$. For SCE, they set $A=-4, \alpha=10, \beta=1$. For NCE+RCE, we set $\alpha=50, \beta=0.1$. For NCE+AGCE, we set $\alpha=50, \beta=0.1, a=2.5, q=3$. For NCE+NNCE we set $\alpha=20, \beta=1$. For NFL+NNFL we set $\alpha=20, \beta=1$.
For VCE, we set $a = 0.015$ and scale by. For NCE+VCE, we set $\alpha=5, \beta=0.5, a=0.03$.

\paragraph{Experiment Setting for Clothing1M.}
We follow the same experiment setting in \citep{ANL}.
For Clothing1M, we use ResNet-50 pre-trained on ImageNet. All the images are resized to $224 \times 224$. We use SGD with a momentum of 0.9, a weight decay of $1 \times 10^{-3}$, and batch size of 32. We train the network for 10 epochs with a learning rate of $1 \times10^{-3}$ and a decay of 0.1 at 5 epochs. Typical data augmentations including random shift and horizontal flip are applied. 


\paragraph{Parameters Setting for Clothing1M.}
For GCE, we set $q=0.6$. . For NCE+RCE, they set $\alpha=10, \beta=1, A=-4$. For NCE+NNCE, we set $\alpha=5, \beta=0.1$.
For VCE, we set $a = 0.05$ and scale by 0.1. For NCE+VCE, we set $\alpha=1, \beta=0.1, a=0.2$.

%% file: figs/ece.tex
\begin{figure*}[!h]
    \centering
    \subfigure[CE]{
    \includegraphics[width=1.55in]{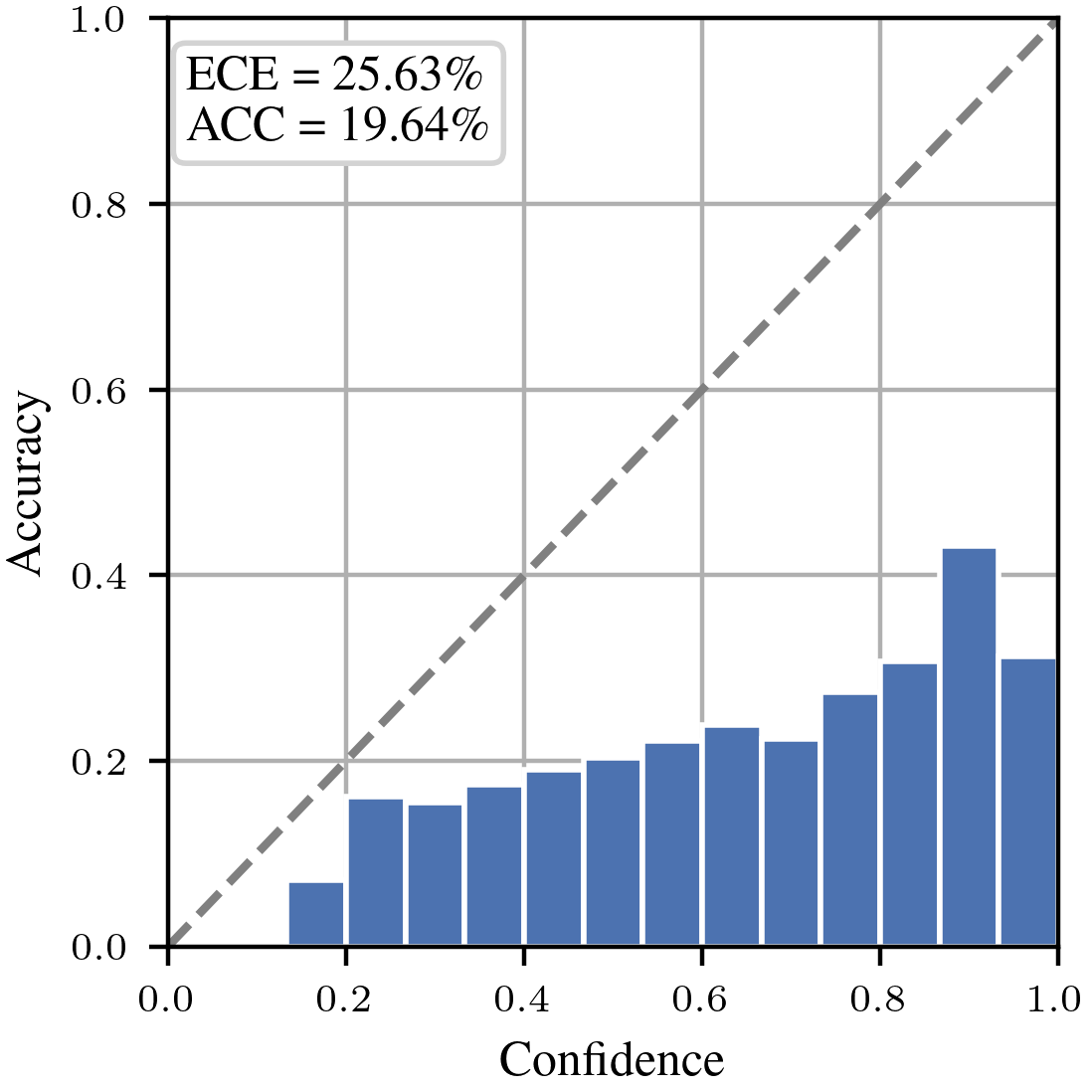}
    }
    \subfigure[VCE]{
    \includegraphics[width=1.55in]{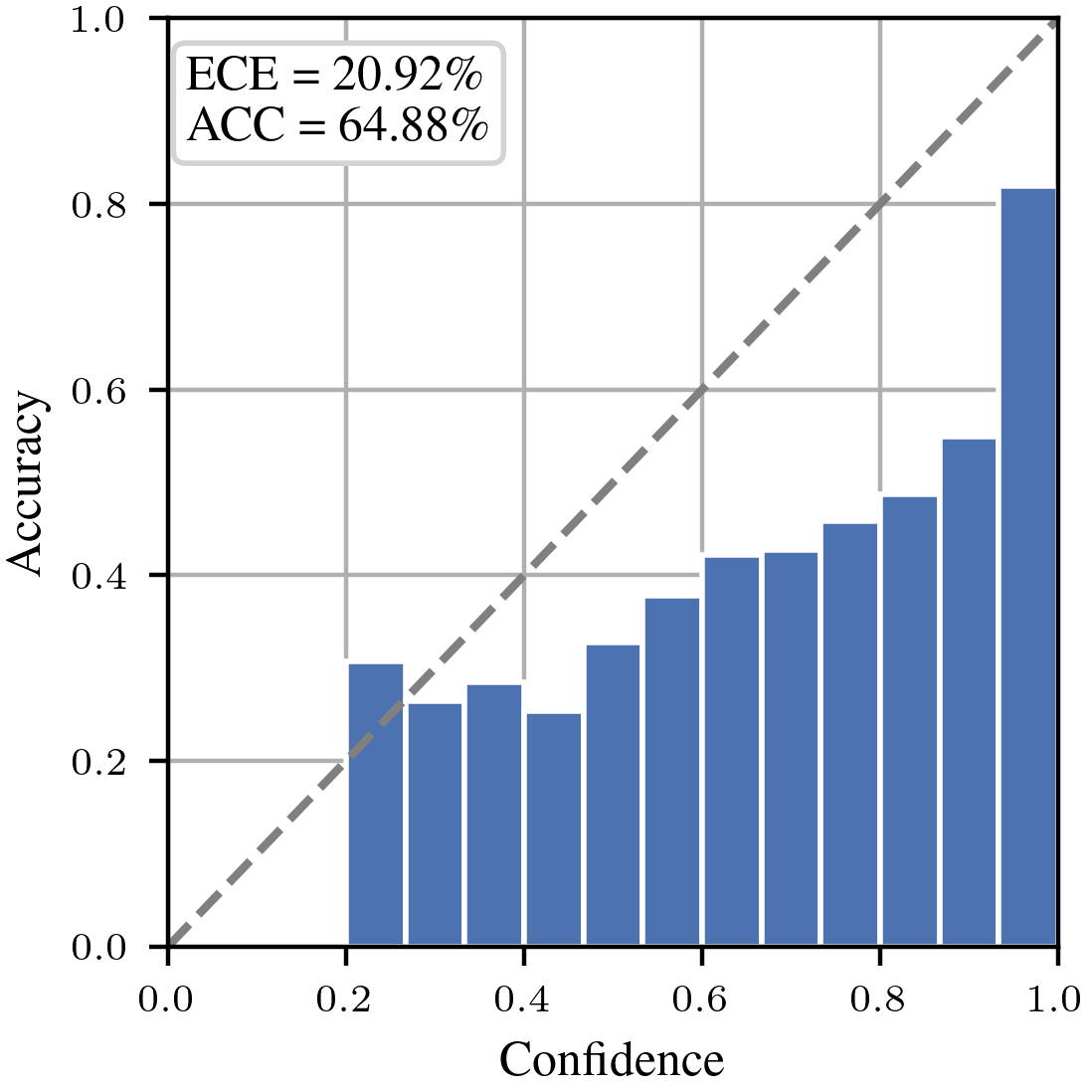}
    } 
    \subfigure[VEL]{
    \includegraphics[width=1.55in]{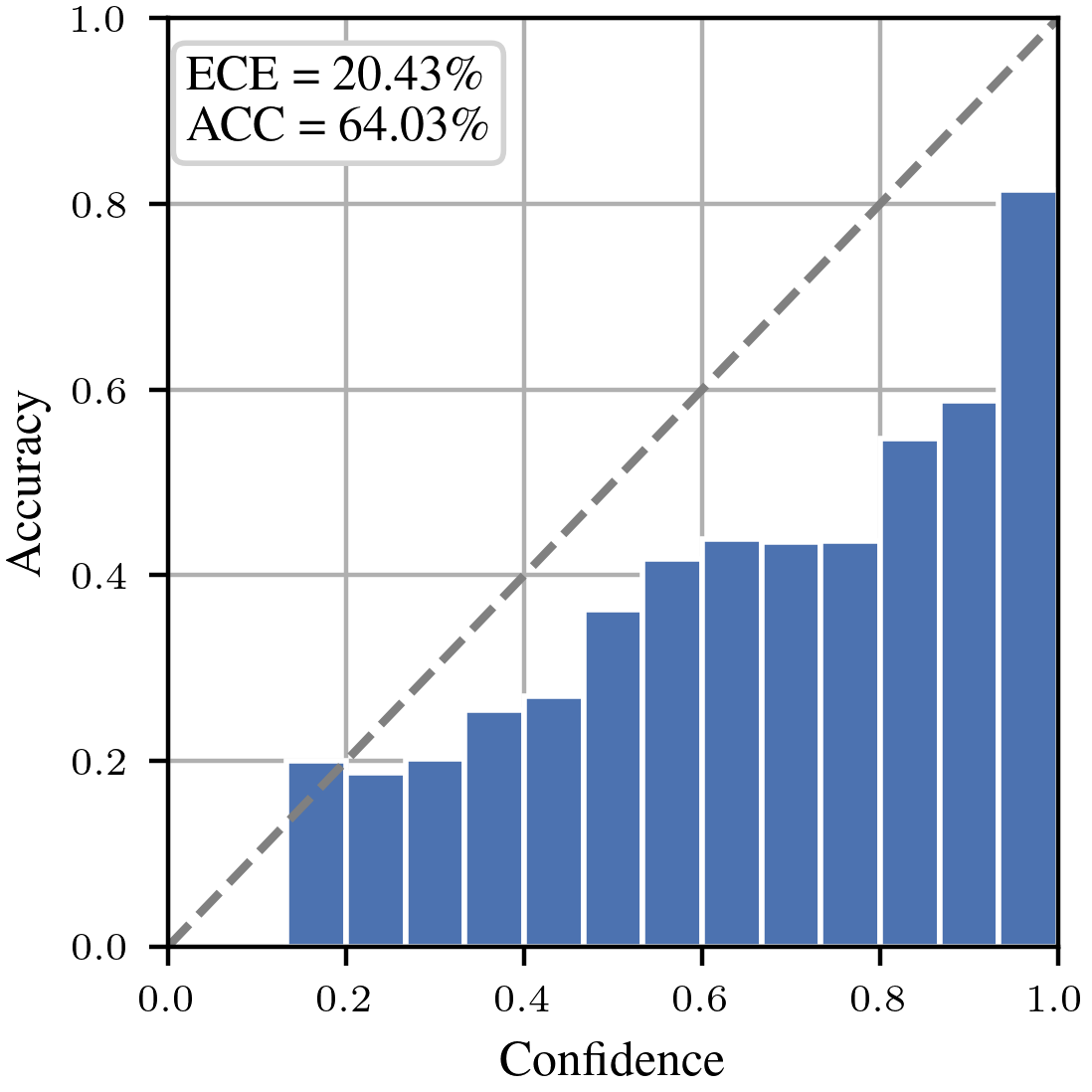}
    }
    \subfigure[VSL]{
    \includegraphics[width=1.55in]{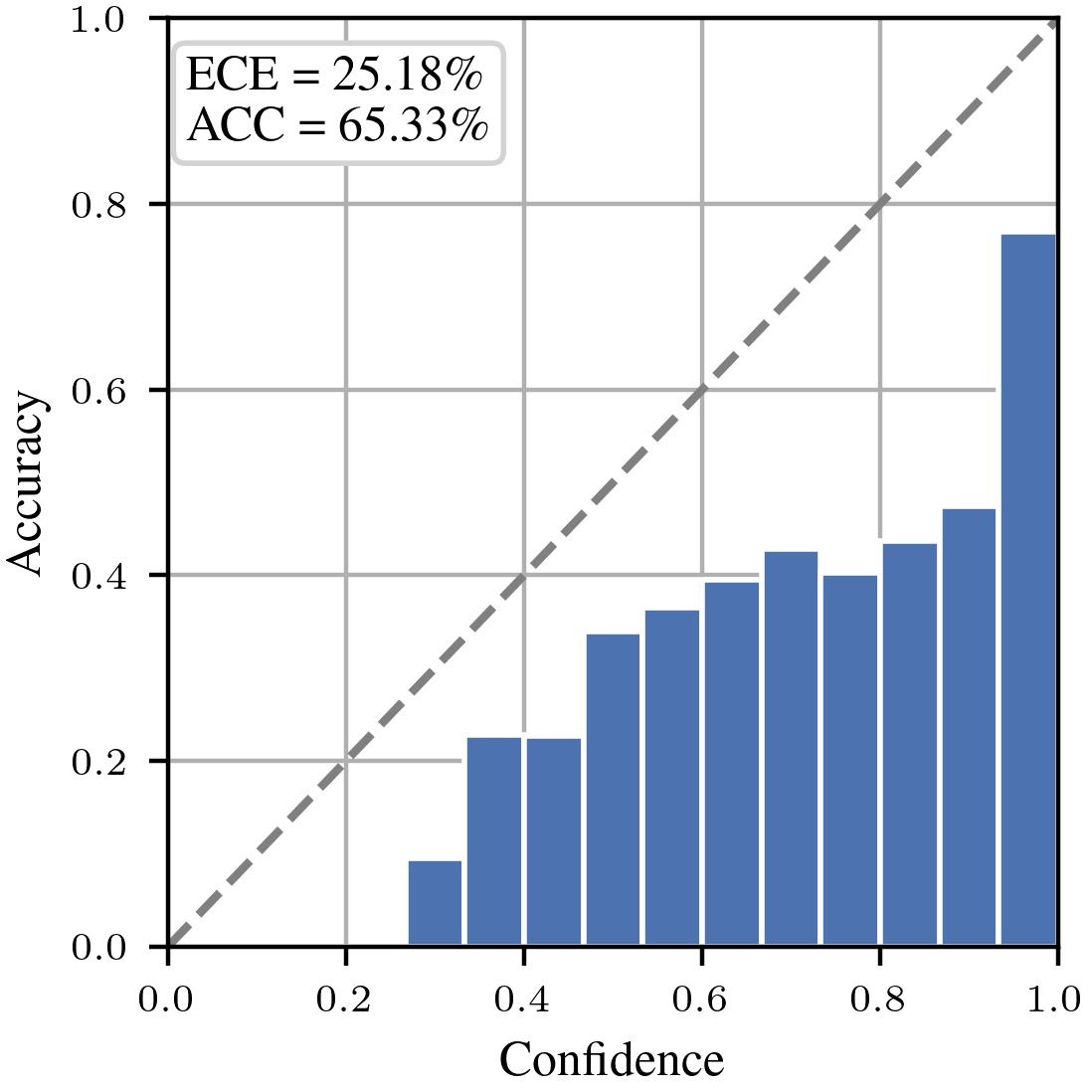}
    }
    \caption{Reliability diagrams of CIFAR-10 with 0.8 symmetric noise. }
    \label{fig:ece}
\end{figure*}